\documentclass[lettersize,journal]{IEEEtran}
\usepackage{amsmath,amsfonts}
\usepackage{algorithm}
\usepackage{array}
\usepackage[caption=false,font=normalsize,labelfont=sf,textfont=sf]{subfig}
\usepackage{textcomp}
\usepackage{stfloats}
\usepackage{url}
\usepackage{verbatim}
\usepackage{graphicx}
\usepackage{cite}
\usepackage{xspace}
\usepackage{authblk}
\usepackage{adjustbox}
\usepackage{multirow}
\usepackage{booktabs}
\usepackage[table,xcdraw]{xcolor}
\usepackage{algpseudocode}
\usepackage{bbding}
\usepackage{orcidlink}
\usepackage{hyperref}

\usepackage{soul}
\soulregister{\cite}{7}
\soulregister{\ref}{7}

\makeatletter
\DeclareRobustCommand\onedot{\futurelet\@let@token\@onedot}
\def\@onedot{\ifx\@let@token.\else.\null\fi\xspace}

\def\etc{\emph{etc}\onedot} 
 
\def\etal{\emph{et al}\onedot}
\makeatother

\begin{document}

\title{Underwater Image Restoration Through a Prior Guided Hybrid Sense Approach and Extensive Benchmark Analysis}

\author{
Xiaojiao Guo, Xuhang Chen\orcidlink{0000-0001-6000-3914}, Shuqiang Wang and Chi-Man Pun
\thanks{This work was supported in part by the University of Macau under Grants MYRG-GRG2023-00131-FST and MYRG-GRG2024-00065-FST-UMDF, in part by the Science and Technology Development Fund, Macau SAR, under Grants 0141/2023/RIA2 and 0193/2023/RIA3, in part by National Natural Science Foundations of China under Grant 62172403 and 12326614.}
\thanks{Xiaojiao Guo, Xuhang Chen and Chi-Man Pun are with the Department of Computer and Information Science, University of Macau, Macao SAR, China (e-mail: yc27441@um.edu.mo; yc17491@umac.mo; cmpun@umac.mo).}
\thanks{Xuhang Chen and Shuqiang Wang are with Shenzhen Institutes of Advanced Technology, Chinese Academy of Sciences, Shenzhen 518055, China (e-mail: xx.chen2@siat.ac.cn; sq.wang@siat.ac.cn).}
\thanks{Xiaojiao Guo is also with the School of Big Data, Baoshan University, Baoshan 678000, China.}
\thanks{Xuhang Chen is also with the School of Computer Science and Engineering, Huizhou University, Huizhou 516007, China (e-mail: xuhangc@hzu.edu.cn).}
\thanks{Xiaojiao Guo and Xuhang Chen contribute equally to this work.}
\thanks{Chi-Man Pun and Shuqiang Wang are the corresponding authors (e-mail: cmpun@umac.mo; sq.wang@siat.ac.cn).}
}

\markboth{Journal of \LaTeX\ Class Files,~Vol.~14, No.~8, August~2021}%
{Shell \MakeLowercase{\textit{et al.}}: A Sample Article Using IEEEtran.cls for IEEE Journals}


\maketitle

\begin{abstract}
Underwater imaging grapples with challenges from light-water interactions, leading to color distortions and reduced clarity. In response to these challenges, we propose a novel Color Balance Prior \textbf{Guided} \textbf{Hyb}rid \textbf{Sens}e \textbf{U}nderwater \textbf{I}mage \textbf{R}estoration framework (\textbf{GuidedHybSensUIR}). This framework operates on multiple scales, employing the proposed \textbf{Detail Restorer} module to restore low-level detailed features at finer scales and utilizing the proposed \textbf{Feature Contextualizer} module to capture long-range contextual relations of high-level general features at a broader scale. The hybridization of these different scales of sensing results effectively addresses color casts and restores blurry details. In order to effectively point out the evolutionary direction for the model, we propose a novel \textbf{Color Balance Prior} as a strong guide in the feature contextualization step and as a weak guide in the final decoding phase. We construct a comprehensive benchmark using paired training data from three real-world underwater datasets and evaluate on six test sets, including three paired and three unpaired, sourced from four real-world underwater datasets. Subsequently, we tested 14 traditional and retrained 23 deep learning existing underwater image restoration methods on this benchmark, obtaining metric results for each approach. This effort aims to furnish a valuable benchmarking dataset for standard basis for comparison. The extensive experiment results demonstrate that our method outperforms 37 other state-of-the-art methods overall on various benchmark datasets and metrics, despite not achieving the best results in certain individual cases. The code and dataset are available at \href{https://github.com/CXH-Research/GuidedHybSensUIR}{https://github.com/CXH-Research/GuidedHybSensUIR}. 
\end{abstract}


\section{Introduction}
\begin{figure}[ht]
\centering
\includegraphics[width=\columnwidth]{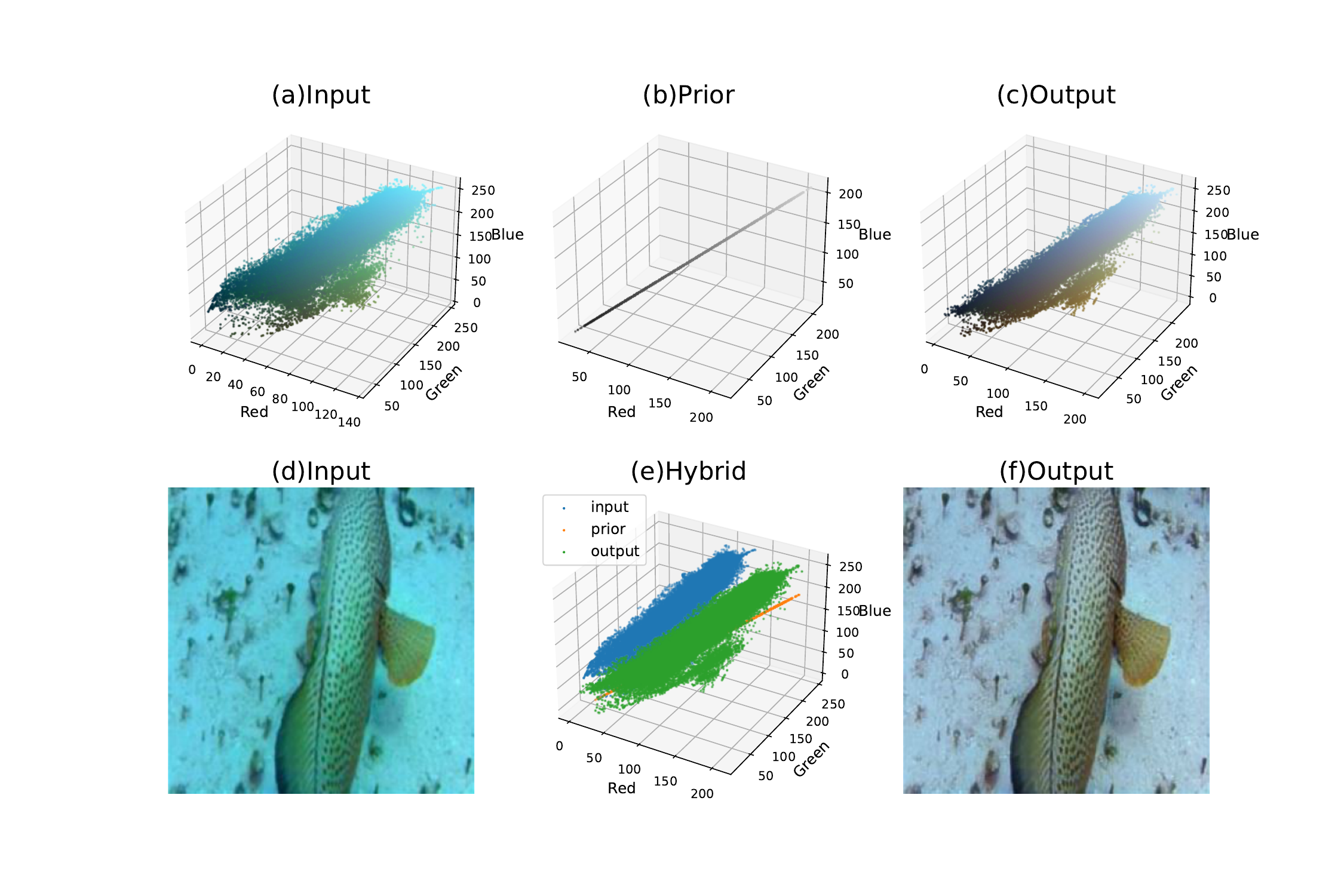}
\caption{Color distribution visualized by 3D scatter plot. (a), (b), and (c) represent the individual color scatter plots of the input underwater image, color balance prior, and the output restored image of our model, respectively. To enhance comparison, these scatter plots are amalgamated in (e). The corresponding input and output images are presented in (d) and (f) for visual reference.}
\label{fig:scatter}
\end{figure}

\IEEEPARstart{T}{he} underwater world is a mysterious realm abundant with lively ecosystems, fascinating geological features, ancient wrecks, and undocumented species. Underwater imaging is crucial for exploring and protecting this environment. However, underwater images often suffer from a dominant blue or green hue due to the differential absorption of light wavelengths in water: red light is absorbed most quickly, followed by green, then blue. Fig. \ref{fig:scatter} (a) visually represents this selective absorption effect. Additionally, blurriness arises from light absorption, scattering, and suspended particles. These challenges degrade image quality, hindering high-level visual tasks such as marine segmentation \cite{li2021marine}, object localization \cite{zhao2018weakly}, and recognition \cite{zhao2019look,tu2021joint}. Therefore, enhancing underwater image quality is essential to improve the accuracy and efficiency of these tasks, advancing our understanding of the underwater world.

Addressing these issues requires going beyond traditional image enhancement methods like correcting biased colors, enhancing saturation, and improving contrast. The unique underwater challenges, including restoring details blurred by low light and turbidity, must also be tackled. Thus, we term our task \textit{Underwater Image Restoration (UIR)}, emphasizing our goal of comprehensively restoring underwater images to in-air quality.

Recently, UIR has seen significant developments from both traditional and deep learning methods. Although traditional methods like ROP \cite{liu2022rank} and WWPF \cite{zhang2023underwater} have made notable advancements, their performance can be inconsistent due to diverse water types and complex environments. In contrast, deep learning methods are gaining prominence thanks to their robust adaptability to challenging underwater conditions. Most deep learning UIR methods have predominantly used CNNs \cite{wang2023domain,wen2023syreanet,huang2023contrastive,xue2023investigating}, which excel at extracting local features but are less effective at modeling long-range dependencies. This limitation prompted the recent incorporation of Transformers in UIR methods \cite{peng2023u}. However, prior UIR approaches have not explored combining CNNs and Transformers to leverage their collective strengths.

To achieve our dual objectives – restoring fine-scale blurry details and enhancing color performance at a coarser scale – we have employed a combination of both CNNs and Transformers to create our ``Hybrid Sense UIR'' network. This hybrid approach leverages the strengths of each, allowing for a more comprehensive and effective restoration of underwater images. Considering our need to handle features ranging from fine to coarse scales, we have selected the U-shape architecture, which is highly suitable for processing multi-scale features. In the encoding phase of this U-shape structure, we've developed a module called ``Detail Restorer'', implemented using CNNs, to concentrate specifically on local information across various scales. Subsequently, preceding the decoding stage, we incorporated a Transformer-based bottleneck module named ``Feature Contextualizer'', designed to focus more comprehensively on global color relationships. To further emphasize the focus on overall color dynamics in the ``Feature Contextualizer'', we have modified the traditional ViT \cite{dosovitskiy2020image} approach. This adaptation involves shifting from the standard practice of computing attention among divided image patches to implementing inter-channel attention. This strategic change prioritizes the nuanced interactions between color channels, thereby offering a more color-centric analysis and interpretation.

To tackle the aforementioned color distortion challenges, we introduce a novel color balance prior to guide our model's color restoration process. 
This prior, inspired by the Gray World Assumption \cite{buchsbaum1980spatial}, is defined as the mean value of the RGB channels for each pixel in the original image. The Gray World Assumption suggests that in a typically-illuminated scene, these average color channel values should be similar, resulting in a neutral gray. While underwater environments may deviate from this ideal, the principle of balanced color channels remains relevant for achieving realistic color restoration.
As shown in Fig. \ref{fig:scatter} (b) and (e), the trajectory of this prior is consistent with the angle of inclination observed when comparing the color distribution between the original and the restored underwater images. Crucially, this prior occupies a central position in the scatter plot of the restored image, indicating that it approximates the average color distribution of our restoration aim. This central position suggests that the color balance prior is well-suited for directing the restoration process toward more accurate results. By leveraging this prior-guided approach in conjunction with our ``Hybrid Sense UIR'' network, we aim to effectively address the challenges of color distortion and enhance the overall quality of restored underwater images.

Our extensive review of recent developments in underwater image restoration and enhancement methods revealed a notable inconsistency: there is a lack of standardization in benchmarks, as almost no two methods employ the same datasets for training and testing, nor do they use identical metrics for evaluation.
For their training datasets, researchers either employed a single dataset, as exemplified by GUPDM \cite{mu2023generalized}, which was solely trained on LSUI \cite{peng2023u}, or more commonly opted for a varied approach by randomly selecting images from 2-3 different sources, such as Ucolor \cite{li2021underwater}, TUDA \cite{wang2023domain}, Semi-UIR \cite{huang2023contrastive} \etc To establish a robust benchmark that advances future research in underwater image restoration, we compiled a comprehensive dataset by aggregating 5600 training images, 490 paired testing data, and 460 test samples without referenced targets from four diverse real-world underwater image datasets: UIEB \cite{li2019underwater}, EUVP \cite{islam2020fast}, LSUI \cite{peng2023u}, and RUIE \cite{liu2020real}. We then carried out an extensive benchmarking exercise involving 37 existing UIR methods. This provides a standardized basis for comparison and yields meaningful insights into performance improvements. 

Overall, our contributions can be summarized as follows:
\begin{itemize}
    
    
    
    \item We propose a novel \textbf{Hybrid Sense UIR Framework} integrating the Detail Restorer, Feature Contextualizer and Scale Harmonizer modules, enabling effective multi-scale feature processing to correct color distortions and restore blurred details.

    \item By introducing a novel \textbf{Color Balance Prior}, we enhance our framework into the \textbf{Guided Hybrid Sense UIR Framework}, significantly improving restoration quality.

    \item We assembled an extensive dataset combining images from four real-world underwater datasets, retrained 37 existing UIR methods, and conducted a thorough comparative performance analysis, establishing a robust benchmark.

    \item Extensive experiments demonstrate that our framework outperforms state-of-the-art methods, delivering superior results in underwater image restoration.
\end{itemize}

\section{Related Work}
\subsection{UIR Methods}
The field of underwater image restoration (UIR) has evolved from traditional approaches to sophisticated deep learning techniques, reflecting the growing challenges in underwater imaging and the demand for higher-quality restorations.
\subsubsection{Traditional Approaches}
Traditional methods in underwater image restoration have largely been divided into two categories: model-based and model-agnostic. 


\textbf{Model-based methods} rely on manually designed priors to determine variables like transmission rates and background illumination in underwater image models. The Dark Channel Prior (DCP) \cite{he2010single}, initially developed for dehazing, has been adapted for underwater restoration in works such as \cite{drews2013transmission, li2016single, peng2018generalization, xie2021variational, zhou2021underwater}. The Underwater Light Attenuation Prior (ULAP) uses intensity analysis across color channels to generate depth maps, aiding in color and contrast correction \cite{chiang2011underwater, akkaynak2019sea, zhou2023underwater}. Other priors include statistical priors \cite{song2020enhancement}, minimum color loss prior \cite{li2016underwater, zhang2022underwater}, blurriness prior \cite{peng2017underwater}, haze lines prior \cite{berman2017diving, berman2020underwater}, rank one prior \cite{liu2022rank}, and illumination channel sparsity prior \cite{hou2023non}. While effective in controlled conditions, these methods often struggle with the complexities of real-world underwater scenes.

\textbf{Model-agnostic methods} operate independently of physical models, utilizing image processing techniques like contrast adjustment, histogram equalization, and color correction. Ancuti \etal \cite{ancuti2012enhancing} developed a single-image solution based on multi-scale fusion principles without relying on specialized optical models like UIFM \cite{jaffe1990computer}. Techniques such as Contrast Limited Adaptive Histogram Equalization (CLAHE) \cite{hitam2013mixture} and the Unsupervised Color Correction Method (UCM) \cite{iqbal2010enhancing} enhance image contrast and visibility effectively. Refinements of the Retinex algorithm have been made for underwater enhancement, integrating bilateral and trilateral filters \cite{zhang2017underwater}, Bayesian approaches with multi-order gradient priors \cite{zhuang2021bayesian}, and hyper-Laplacian reflectance priors \cite{zhuang2022underwater}. Multiscale fusion strategies have been employed by Ancuti \etal \cite{ancuti2017color} and Jiang \etal \cite{kang2022perception} to blend images from color-compensated and white-balanced versions, enhancing global contrast and edge sharpness. However, these methods may produce inconsistent results due to inadequate consideration of varying degrees of degradation in underwater environments, potentially leading to over- or under-enhanced image regions.

\subsubsection{Deep Learning-based Approaches}
Early deep learning UIR techniques \cite{wang2019underwater, kar2021zero} primarily relied on physical imaging models, using neural networks to predict elements like transmission and ambient light. Similar to traditional models, they often struggled with the unpredictability of real-world underwater scenes.

Recent approaches shift towards learning directly from paired datasets of underwater images and clear references, eliminating reliance on imaging models. Convolutional Neural Networks (CNNs) are commonly employed, with models such as WaterNet \cite{li2019underwater}, UWCNN \cite{li2020underwater}, Ucolor \cite{li2021underwater}, CLUIE-Net \cite{li2022beyond}, Shallow-UWNet \cite{naik2021shallow}, PUIE-Net \cite{fu2022uncertainty}, STSC \cite{wang2022semantic}, DeepWaveNet \cite{sharma2023wavelength}, $\mathrm{NU^2Net}$ \cite{guo2023underwater}, and SFGNet \cite{zhao2024toward}.

GAN-based models have also been explored extensively, starting with the adaptation of CycleGAN \cite{zhu2017unpaired} by Fabbri \etal \cite{fabbri2018enhancing}. Subsequent frameworks include UGAN \cite{fabbri2018enhancing}, CWR \cite{han2021single}, TACL \cite{liu2022twin}, UIE-WD \cite{ma2022wavelet}, PUGAN \cite{cong2023pugan}, and TUDA \cite{wang2023domain}. Despite their potential, GANs face challenges like model collapse and training instability, requiring substantial computational resources.

To leverage the Transformer's ability to capture long-range dependencies, architectures like the U-shaped Transformer \cite{peng2023u} and a U-Net-based reinforced Swin-Convs Transformer \cite{ren2022reinforced} have been introduced. Addressing the scarcity of labeled data, semi-supervised (e.g., Semi-UIR \cite{huang2023contrastive}) and unsupervised learning approaches (e.g., USUIR \cite{fu2022unsupervised}) have been adopted, enhancing model learning efficiency and accuracy in data-scarce environments.

Deep learning models integrated with priors are also prevalent. Ucolor \cite{li2021underwater} uses a medium transmission-guided multi-color space embedding to address color casts and low contrast due to wavelength and attenuation. Mu \etal \cite{mu2023generalized} proposed a dynamic, physical-knowledge-guided method for adaptive enhancement. CCMSRNet \cite{qi2023deep} incorporates illumination estimations from a Multiscale Retinex Network to improve visibility. The underwater image quality assessment method URanker \cite{guo2023underwater} utilizes color histograms as priors to address global degradation.

Despite advancements, our quantitative and qualitative analyses indicate that these deep learning methods still have room for improvement.

\subsection{Underwater Image Datasets}
Underwater image datasets are typically categorized into synthetic and real-world datasets.

Due to the difficulty of obtaining clear images corresponding directly to underwater photos, some models, including WaterGAN \cite{li2017watergan} and UGAN \cite{fabbri2018enhancing}, use Generative Adversarial Networks (GANs) to synthesize underwater images for training. Li \etal \cite{li2020underwater} synthesized a dataset comprising 10 subsets for different water types and degradation levels. However, a gap persists between synthetic and real-world images; models trained on synthetic data often struggle in real scenes. Consequently, we exclude synthetic datasets from our benchmark to ensure relevance to real-world applications.

Real-world datasets are crucial for developing and evaluating UIR methods and are divided into paired and non-reference datasets.
\begin{figure*}[ht]
\centering
\includegraphics[width=0.8\textwidth]{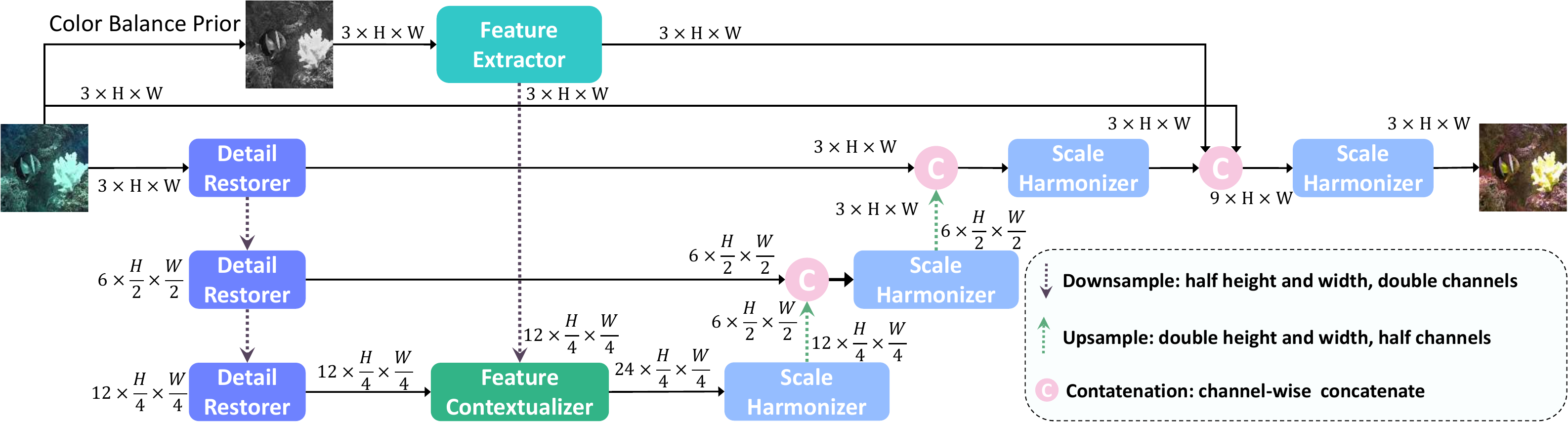}
\caption{Overview of the GuidedHybSensUIR architecture—a U-shaped model with encoder, bottleneck, and decoder stages enabling hybrid sense-level feature processing. Detail Restorer modules in the decoder enhance local details at three scales. At the bottleneck, the Feature Contextualizer—guided by the color balance prior—captures long-range relationships and global color dependencies for effective color correction and enhancement. Scale Harmonizer modules in the decoder refine fused information after each fusion step, ensuring seamless multi-scale integration for high-quality image reconstruction.}
\label{fig:overall}
\end{figure*}

\textbf{Paired datasets} include both underwater images and corresponding reference images. The UIEB dataset \cite{li2019underwater} contains 950 images, with 890 paired references. The EUVP dataset \cite{islam2020fast} features over 11,000 paired and more than 8,000 unpaired images. The LSUI dataset \cite{peng2023u} offers over 4,000 paired images across diverse underwater scenes. References in these datasets are selected using ensemble learning principles \cite{dietterich2002ensemble} to combine strengths from multiple UIR methods. Specifically, in the EUVP dataset, the references are sampled from a large set of real underwater images; images with minimal distortion are selected, and then CycleGAN is trained to generate the distorted counterparts. While some references may not be optimal, these datasets remain vital for advancing deep learning methods in UIR.

\textbf{Non-reference datasets}, lacking corresponding reference images, still significantly contribute by providing diverse real-world images for testing and evaluation. The RUIE dataset \cite{liu2020real} comprises over 4,000 images with diverse illumination, depth of field, blurring, and color casts. The SQUID dataset \cite{akkaynak2019sea} includes over 1,100 images from two optically distinct water bodies under natural illumination.

The availability of these diverse datasets is essential for developing robust UIR methods. By encompassing various underwater environments, water types, and degradation levels, these datasets ensure that trained models can generalize effectively to real-world scenarios.

\section{Methodology}

Our proposed ``GuidedHybSensUIR'' architecture, depicted in Fig. \ref{fig:overall}, employs a U-shaped network to address underwater image restoration from both local and global perspectives.

For global enhancement, we propose a Transformer-based Feature Contextualizer at the bottleneck to model long-range relationships and capture global color dependencies, transforming dark underwater scenes into distortion-free appearances. For local restoration, we use CNN-based Detail Restorer modules within the encoder to recover textural details and enhance image sharpness. The decoder merges bottleneck and encoder features, combining global context with fine-grained details. Scale Harmonizer modules refine this fusion after each step, ensuring adaptive integration across scales for high-quality reconstruction.

Furthermore, to improve color balance and visual quality, we formulate the color balance prior into the Feature Contextualizer and the final Scale Harmonizer, upgrading the model to ``Guided Hybrid Sense UIR''. This prior provides strong guidance for global color correction and weak guidance for color updates during decoding.




\subsection{Hybrid Sense Architecture}
The architecture of our proposed model utilizes a U-Net-like structure. Initially, when an underwater image $X$ of dimensions $C \times H \times W$ is inputted, it is first processed by a Detail Restorer designed to encode its local details. Following this, the output is downsampled to a coarser scale with dimensions $2C \times \frac{H}{2} \times \frac{W}{2}$, and the Detail Restoration process is repeated. It is important to note that when downsampling the features from the previous scale to half its spatial dimensions, we double the number of channels. This approach ensures that there is no loss of information during the downsampling process.

\subsubsection{Detail Restorer}
In our Detail Restorer, the basic unit is a quaternion convolution structure, which integrates a Residual Context Block (RCB) and a Nonlinear Activation-Free Block (NAFB), the latter is specifically inspired by the design of NAFNet \cite{chen2022simple}. The network comprises six sequential quaternion convolutional units.

In the quaternion CNNs developed by Zhu \etal \cite{zhu2018quaternion}, the R, G, and B channels are represented as the three separate imaginary components of a quaternion, with the real component set to zero. This representation allows the network to capture interdependencies among the color channels, leading to more representative features. 

Building upon this idea, we utilize quaternion representation to integrate the outputs of our parallel branches — the RCB and the NAFB — into a unified framework. In common approaches, integrating information from different branches often involves directly concatenating or adding their outputs before the next operation. However, such methods may introduce unnecessary degrees of freedom in the fusion process, potentially leading to instability and less effective cooperation between branches.

By employing quaternion convolution, we limit the degrees of freedom in the fusion of the parallel branches. The quaternion algebra imposes mathematical constraints on how the feature maps from the RCB and NAFB are combined. This structured fusion fosters more stable cooperation between these two parallel blocks, enhancing the network's ability to capture intricate features necessary for detail restoration.

Let $\mathbf{x}$ be the input tensor to both the RCB and NAFB. The outputs of these blocks are denoted as:
\begin{align}
{\small
\mathbf{A} = \text{RCB}(\mathbf{x}), 
\mathbf{B} = \text{NAFB}(\mathbf{x}),
}
\end{align}
where $\mathbf{A}$ and $\mathbf{B}$ are real-valued tensors representing feature maps of the same dimensions as $\mathbf{x}$.

We define the quaternion feature $\mathbf{Q}$ as:
\begin{equation}
{\small
\mathbf{Q} = \mathbf{R} \cdot 1 + \mathbf{A} \cdot i + \mathbf{B} \cdot j + \mathbf{C} \cdot k,
}
\label{eq:quaternion_feature}
\end{equation}
where $\mathbf{R}=\mathbf{0}$ and $\mathbf{C}=\mathbf{0}$ are tensors set to zeros of the same dimensions as $\mathbf{A}$ and $\mathbf{B}$. Here, $1$, $i$, $j$, and $k$ are the quaternion units satisfying the Hamilton product rules ($i^2 = j^2 = k^2 = ijk = -1$).

The quaternion convolution is performed by applying a quaternion-valued convolutional kernel $\mathbf{W}$ to $\mathbf{Q}$:
\begin{equation}
{\small
\mathbf{Q}_{\text{out}} = \mathbf{Q} * \mathbf{W},
}
\end{equation}
where $\mathbf{W} = \mathbf{W}_r 1 + \mathbf{W}_i i + \mathbf{W}_j j + \mathbf{W}_k k$ and $*$ denotes the Hamilton product.

Given that $\mathbf{R} = \mathbf{0}$ and $\mathbf{C} = \mathbf{0}$, after expanding the Hamilton product with our simplified quaternion input, we have:
\begin{equation}
{\small
\begin{split}
&\mathbf{Q}_{\text{out}, r} = -\mathbf{A}\mathbf{W}_i - \mathbf{B}\mathbf{W}j, \\
&\mathbf{Q}_{\text{out}, i} = \mathbf{A}\mathbf{W}_r - \mathbf{B}\mathbf{W}k, \\
&\mathbf{Q}_{\text{out}, j} = \mathbf{A}\mathbf{W}_k + \mathbf{B}\mathbf{W}r, \\
&\mathbf{Q}_{\text{out}, k} = \mathbf{A}\mathbf{W}_j - \mathbf{B}\mathbf{W}_i.
\end{split}
}
\end{equation}

This convolution captures the interactions between the feature maps from the RCB and NAFB in a mathematically constrained manner, integrating them into a cohesive representation. By limiting the degrees of freedom in the fusion process through quaternion algebra, we foster a more stable and cooperative fusion of features compared to direct concatenation or addition.

\begin{figure}[ht]
\centering
\includegraphics[width=\columnwidth]{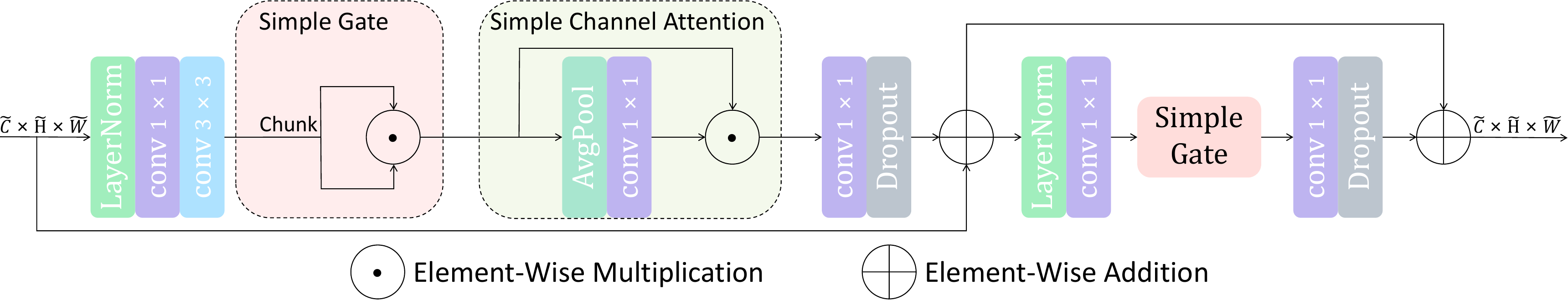}
\caption{Architecture of the Nonlinear Activation-Free Block (NAFB).}
\label{fig:NAFB}
\end{figure}

By employing simplified gating mechanisms, the NAFB selectively passes or suppresses information based on element-wise operations, enhancing the network's ability to restore intricate details. As depicted in Fig. \ref{fig:NAFB}, the main components of the NAFB include two groups of residual blocks. In the first part, a $1 \times 1$ convolution followed by a $3 \times 3$ convolution is applied to the input features. Subsequently, a simple gate splits the resulting feature maps into two chunks, which are then multiplied element-wise to achieve pixel-wise self-attention. This operation effectively enhances or suppresses pixel-wise information in a straightforward manner. Next, a simple channel attention mechanism employing global average pooling is applied to calibrate the feature amplitudes by multiplying the attention weights with the feature maps. A $1 \times 1$ convolution and a dropout operation are then conducted, acting as a nonlinear activation. In the second part, we repeat the same operations without the simple channel attention module. Both parts include residual connections from their respective inputs. This technique preserves more useful information across network layers, potentially boosting performance in the detail restoration process.

\begin{algorithm}
        \caption{{\footnotesize RCB (Residual Context Block)}}
        \label{alg_rcb}
{\footnotesize    
        \begin{algorithmic}[1]
            \Require $\mathbf{x}$ (input image features)
            \Ensure $\mathbf{out}$ (output image features)

            \State Apply depth-wise convolution: $\mathbf{r} \gets DwConv2d_{3 \times 3}(\mathbf{x})$
            \State Apply activation: $\mathbf{r} \gets LeakyReLU_{0.2}(\mathbf{r})$
            \State Apply depth-wise convolution: $\mathbf{r} \gets DwConv2d_{3 \times 3}(\mathbf{r})$
            \State Pass $\mathbf{r}$ through the ContextBlock module to obtain the context-aware features
            \State Apply activation: $\mathbf{r} \gets LeakyReLU_{0.2}(\mathbf{r})$
            \State Add the residual $\mathbf{r}$ to the input $\mathbf{x}$: $\mathbf{out} \gets \mathbf{r} + \mathbf{x}$
            \State Return the final output $\mathbf{out}$
        \end{algorithmic}
        \hrule
        \begin{algorithmic}[1]        
            \Statex \textbf{def} ContextBlock:
            \Statex \textbf{Input}: $\mathbf{x}$ of shape $[N, C, H, W]$
            \State $\mathbf{input\_x} \gets \mathbf{x}.view(N, C, HW)$
            \State Unsqueeze $\mathbf{input\_x}$: $[N, C, HW] \rightarrow [N, 1, C, HW]$
            \State Obtain context mask: $\mathbf{mask} \gets Conv2d_{1 \times 1, C \rightarrow 1}(\mathbf{x})$
            \State Reshape $\mathbf{mask}$: $[N, 1, H, W] \rightarrow [N, 1, HW]$
            \State Apply softmax: $\mathbf{mask} \gets Softmax(\mathbf{mask})$
            \State Unsqueeze $\mathbf{mask}$: $[N, 1, HW] \rightarrow [N, 1, HW, 1]$
            \State Compute context by matrix multiplication:
            \Statex $\qquad\qquad \mathbf{context} \gets matmul(\mathbf{input\_x}, \mathbf{mask})$
            \State Reshape $\mathbf{context}$: $[N, 1, C, 1] \rightarrow [N, C, 1, 1]$
            \State Apply convolution: $\mathbf{context} \gets Conv2d_{1 \times 1}(\mathbf{context})$
            \State Apply activation: $\mathbf{context} \gets LeakyReLU_{0.2}(\mathbf{context})$
            \State Apply convolution: $\mathbf{context} \gets Conv2d_{1 \times 1}(\mathbf{context})$
            \State Add channel-wise context to input: $\mathbf{caf} \gets \mathbf{x} + \mathbf{context}$
            \State Return the context-aware features $\mathbf{caf}$
        \end{algorithmic}  
}
\end{algorithm}

\begin{figure}[ht]
\centering
\includegraphics[width=\columnwidth]{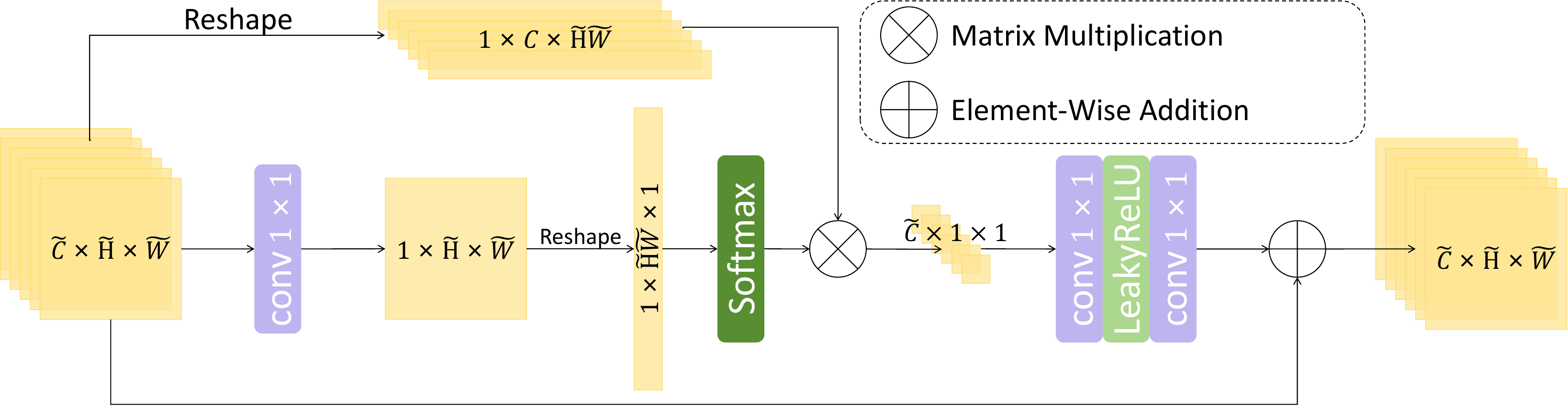}
\caption{Illustration of the ContextBlock within the Residual Context Block (RCB).}
\label{fig:RCB_contextblock}
\end{figure}

Parallel to the NAFB, we incorporate the RCB, as shown in Algorithm \ref{alg_rcb}, which strategically includes a context block. The core component of the RCB is the context block, which focuses on capturing essential contextual information across both channel and spatial dimensions. To elucidate this process, we depict the context block in Fig. \ref{fig:RCB_contextblock}. The initial $1 \times 1$ convolution captures pixel-wise inter-channel contextual information. The resulting contextual feature maps are reshaped and, through matrix multiplication with the reshaped input features, yield a contextual feature for each channel. Finally, after undergoing a group of $1 \times 1$ linear transformation and activation, these contextual features are added to the inputs by a residual connection. This context block is specifically designed to capture essential contextual information, playing a vital assistant role in complementing the NAFB's capabilities in detail restoration.

In short, our Detail Restorer capitalizes on the strengths of quaternion convolutions for handling complex, multidimensional data, the efficiency of NAFBs, and the contextual awareness of RCBs.

\subsubsection{Feature Contextualizer}
\begin{figure}[ht]
\centering
\includegraphics[width=\columnwidth]{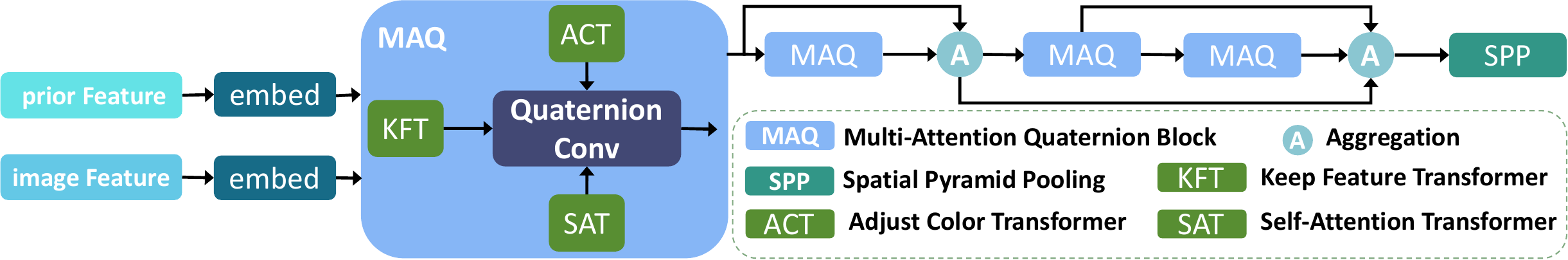}
\caption{Architecture of the Feature Contextualizer module. The $3\times3$ convolutional ``embed'' block projects the $12$ input channels to $48$ embedding channels. The module contains four Multi-Attention Quaternion (MAQ) blocks, each composed of three types of inter-channel attention transformers: the Adjust Color Transformer (ACT), the Keep Feature Transformer (KFT), and the Self-Attention Transformer (SAT). The parallel outputs of these transformers are fused using quaternion convolution.}
\label{fig:FeatureContextualizer}
\end{figure}
After multiple iterations of the Detail Restorer and downsampling processes, where our optimal practice involves three cycles, the features are then fed into the Feature Contextualizer. It is designed to concentrate on contextual information at global scale. The fundamental building block of the Feature Contextualizer is the Multi-Attention Quaternion (MAQ) block. As illustrated in Fig. \ref{fig:FeatureContextualizer}, the Feature Contextualizer consists of a sequence of four MAQ blocks, with residual connections between them. Each MAQ block incorporates three types of Transformers that focus on different aspects of attention: the Adjust Color Transformer (ACT), the Keep Feature Transformer (KFT), and the Self-Attention Transformer (SAT). After the four MAQ blocks, a Spatial Pyramid Pooling (SPP) module \cite{he2015spatial} is employed to calibrate the contextual outputs at four granularities using four levels of pooling and convolution operations.
\begin{figure}[ht]
\centering
\includegraphics[width=\columnwidth]{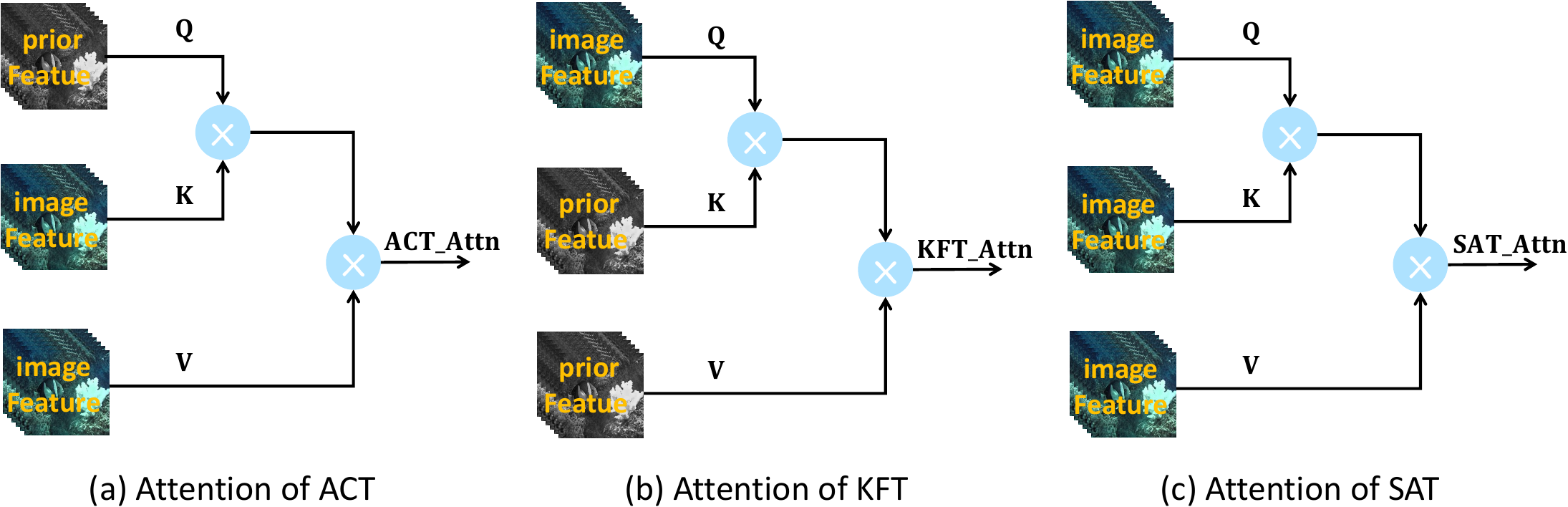}
\caption{Comparison of three different attention mechanisms used in MAQ's Transformers. This diagram focuses on the selection of Query (Q), Key (K), and Value (V) for each attention type, highlighting their distinct attention focus. The calculation details, such as reshape and transpose operations, are omitted for clarity.}
\label{fig:attn}
\end{figure}

The ACT and KFT are designed to capture cross-attention between the image features and the color balance prior features. By attending to the color balance prior, these Transformers enable the network to adapt and refine the color information based on the contextual cues present in the prior features. In contrast, the SAT Transformer focuses on self-attention within the image features themselves, enabling the network to capture and leverage internal relationships and dependencies. Fig. \ref{fig:attn} provides an intuitive understanding of the different attention mechanisms employed by the ACT, KFT, and SAT transformers.

\begin{figure}[ht]
\centering
\includegraphics[width=\columnwidth]{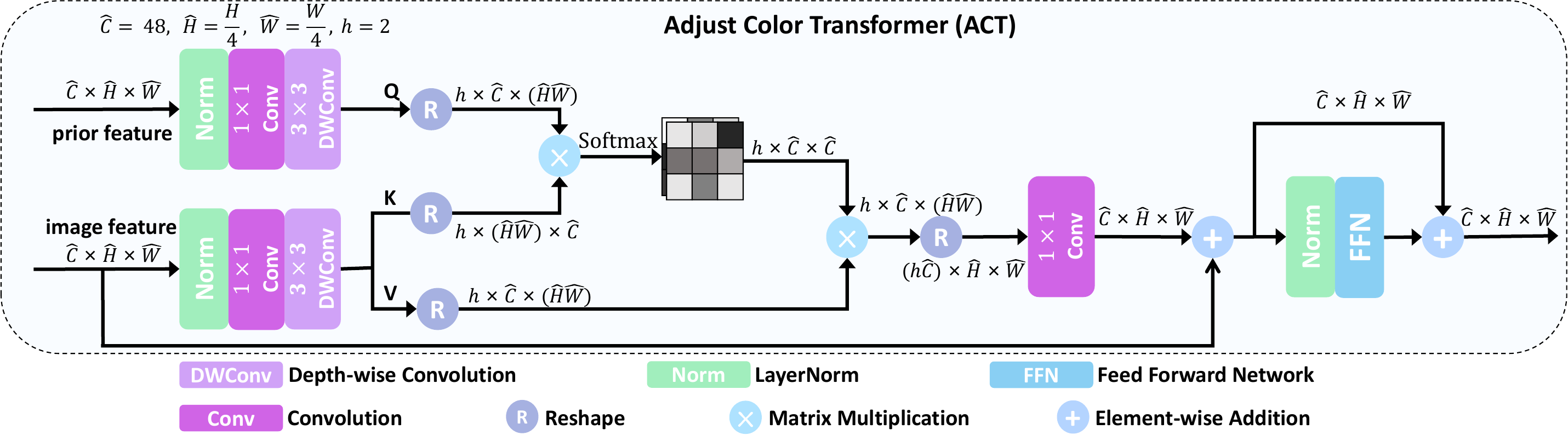}
\caption{Architecture of the Adjust Color Transformer (ACT). The ACT is designed to emphasize the feature channels of the image that have greater similarity to the prior features. To achieve this, the feature maps of the Color Balance Prior are used as the Query (Q), while the feature maps of the input image from the previous step serve as the Key (K) and Value (V). }
\label{fig:ACT}
\end{figure}

The three Transformers within each MAQ block operate in a parallel manner, similar to the Quaternion Block in the Detail Restorer. To achieve this, the real part of the Quaternion Block is set to zero, while the three imaginary parts are assigned to ACT, KFT, and SAT, respectively. By employing this quaternion representation, the degrees of freedom in the parallel branches are constrained, promoting a more balanced and harmonious interaction among different attentions.

Notably, the computational complexity of traditional Vision Transformers (ViTs) is quadratic with respect to the number of image patches, leading to potentially high computational and memory demands. In our earlier Detail Restorer modules, we concentrated on local and detailed features. This emphasis on finer details enables the Feature Contextualizer to shift its focus towards broader aspects, such as inter-channel relationships, rather than inter-patch relationships. Consequently, inspired by the efficient transformer model in Restormer \cite{zamir2022restormer}, the ACT, KFT, and SAT in our approach are tailored to compute attention maps across feature channels. Therefore, the computational complexity becomes linear with respect to the spatial size of the feature maps.
\begin{algorithm}
        \caption{ ACT, KFT and SAT}
        \label{alg_transformer}
{\footnotesize       
        \begin{algorithmic}[1] 
            \Require $\mathbf{prior}$ (prior feature), $\mathbf{x}$ (image feature)
            \Statex $\mathrm{ACT} \leftarrow \mathrm{Attention}(\mathbf{q} \leftarrow \mathbf{prior}, \mathbf{k} \leftarrow \mathbf{x})$
            \Statex $\mathrm{KFT} \leftarrow \mathrm{Attention}(\mathbf{q} \leftarrow \mathbf{x}, \mathbf{k} \leftarrow \mathbf{prior})$
            \Statex $\mathrm{SAT} \leftarrow \mathrm{Attention}(\mathbf{q} \leftarrow \mathbf{x}, \mathbf{k} \leftarrow \mathbf{x})$
        \end{algorithmic} 
        \hrule
        \begin{algorithmic}[1]
            \Statex {\normalsize \textbf{def} Attention:}
            \Statex \textbf{Input:} $\mathbf{q}$ of shape $[N, \hat{C}, \hat{H}, \hat{W}]$, $\mathbf{k}$ of shape $[N, \hat{C}, \hat{H}, \hat{W}]$
            \Statex \textbf{Ouput:} $\mathbf{out}$ of shape $[N, \hat{C}, \hat{H}, \hat{W}]$ (output image feature)
            \State Apply layer normalization: $\mathbf{Q} \gets Norm(\mathbf{q}), \mathbf{K} \gets Norm(\mathbf{k})$
            \State Apply convolution: $\mathbf{Q} \gets Conv2d_{1 \times 1,\hat{C}\rightarrow h\hat{C}}(\mathbf{Q})$
            \Statex $\qquad\qquad\qquad\qquad\; \mathbf{K} \gets Conv2d_{1 \times 1,\hat{C}\rightarrow 2h\hat{C}}(\mathbf{K})$
            \State Split $\mathbf{K}$ into two chunks: $\mathbf{K},\mathbf{V} \gets \mathbf{K}.chunk()$
            \State Apply depth-wise convolution: $\mathbf{Q} \gets DwConv2d_{3 \times 3}(\mathbf{Q})$
            \Statex $\qquad\qquad\qquad\qquad\qquad\qquad\quad \mathbf{K} \gets DwConv2d_{3 \times 3}(\mathbf{K})$
            \Statex $\qquad\qquad\qquad\qquad\qquad\qquad\quad \mathbf{V} \gets DwConv2d_{3 \times 3}(\mathbf{V})$            
            \State Reshape $\mathbf{Q},\mathbf{K},\mathbf{V}$ to get $\mathbf{\hat{Q}},\mathbf{\hat{K}},\mathbf{\hat{V}}$: $[N, h\hat{C}, \hat{H}, \hat{W}] \rightarrow [N, h, \hat{C}, \hat{H}\hat{W}]$
            \State Transpose $\mathbf{\hat{K}}$ to get $\mathbf{\hat{K}}^\top$: $[N, h, \hat{C}, \hat{H}\hat{W}] \rightarrow [N, h, \hat{H}\hat{W}, \hat{C}]$
            \State Calculate attention: $\mathbf{attn} \gets (\hat{\mathbf{Q}} \cdot \hat{\mathbf{K}}^\top) / \tau$
            \State Apply softmax: $\mathbf{attn} \gets Softmax(\mathbf{attn})$
            \State Apply attention: $\mathbf{attn\_out} \gets \mathbf{attn} \cdot \hat{\mathbf{V}}$
            \State Reshape $\mathbf{attn\_out}$: $[N, h, \hat{C}, \hat{H}\hat{W}] \rightarrow [N, h\hat{C}, \hat{H}, \hat{W}]$   
            \State Projection: $\mathbf{attn\_out} \gets Conv2d_{1 \times 1,h\hat{C}\rightarrow \hat{C}}(\mathbf{attn\_out})$
            \State Residual adding: $\mathbf{out} \gets \mathbf{attn\_out} + \mathbf{k}$
            \State Apply layer normalization: $\mathbf{out} \gets Norm(\mathbf{out})$
            \State Go thought Feed Forward Network: $\mathbf{out} \gets FFN(\mathbf{out})$
            \State Residual adding: $\mathbf{out} \gets \mathbf{out} + \mathbf{attn\_out}$
            \State Return $\mathbf{out}$ as the output image feature
        \end{algorithmic} 
}
\end{algorithm}

Specifically, the inputs of the ACT and KFT are the image features and the color balance prior features, both with dimensions $\hat{C} \times \frac{H}{4} \times \frac{W}{4}$, where $\hat{C}$ is the embedding dimension of the features. For simplicity, we denote $\hat{H} = \frac{H}{4}$ and $\hat{W} = \frac{W}{4}$.

The detailed architecture of the Adjust Color Transformer (ACT) is shown in Fig. \ref{fig:ACT}. The goal of ACT is to adjust the color of the image to make it more consistent with the color of that in-air, with the help of the color balance prior. To achieve this, we attend to those feature channels of the image features that have more similarity to the prior features. Therefore, we calculate the inter-channel cross-attention between the prior feature and the image feature, and then multiply this attention to the image feature to finally adjust its contextual information.

As shown in Fig.\ \ref{fig:ACT} and Algorithm \ref{alg_transformer}, we first apply Layer Normalization to both inputs to establish a standard basis for computation. We then use $1 \times 1$ convolutions to expand the prior channels to $h \times \hat{C}$ and the image channels to $2h \times \hat{C}$, where $h$ is the number of attention heads. The prior feature becomes the query $\mathbf{Q} \in \mathbb{R}^{h\hat{C} \times \hat{H} \times \hat{W}}$, and the image feature is split into key $\mathbf{K} \in \mathbb{R}^{h\hat{C} \times \hat{H} \times \hat{W}}$ and value $\mathbf{V} \in \mathbb{R}^{h\hat{C} \times \hat{H} \times \hat{W}}$. All undergo a $3 \times 3$ depth-wise convolution to encode spatial context.

Subsequently, we flatten the last two dimensions of $\mathbf{Q}$, $\mathbf{K}$, and $\mathbf{V}$. The attention map $\mathbf{attn} \in \mathbb{R}^{h \times \hat{C} \times \hat{C}}$ is computed as the dot product of $\mathbf{Q}$ and $\mathbf{K}^\top$, and the output is obtained by multiplying the softmax-operated $\mathbf{attn}$ with $\mathbf{V}$. The entire inter-channel attention operation is succinctly defined as:
\begin{equation}
{\small
\text{Inter\_C\_Attn}(\mathbf{Q}, \mathbf{K}, \mathbf{V}) = \text{Softmax}\left(\frac{\hat{\mathbf{Q}} \cdot \hat{\mathbf{K}}^\top}{\tau}\right) \cdot \hat{\mathbf{V}},
}
\label{eq:inter_c_attn}
\end{equation}
where $\hat{\mathbf{Q}} \in \mathbb{R}^{h \times \hat{C} \times \hat{H}\hat{W} }$, $\hat{\mathbf{K}}^\top \in \mathbb{R}^{h \times \hat{H}\hat{W} \times \hat{C} }$ and $\hat{\mathbf{V}} \in \mathbb{R}^{h \times \hat{C} \times \hat{H}\hat{W} }$ are derived by reshaping tensors from their original dimensions $ h\hat{C} \times \hat{H} \times \hat{W}$. Here, $\tau$ acts as a learnable temperature parameter that modulates the magnitude of the dot product between $\mathbf{Q}$ and $\mathbf{K}$, analogous to the role it plays in conventional self-attention mechanisms.

To more clearly demonstrate the relationship between the input features and the attention mechanism, the Query ($\mathbf{Q}$), Key ($\mathbf{K}$), and Value ($\mathbf{V}$) in Eq. (\ref{eq:inter_c_attn}), as shown in Fig. \ref{fig:attn}(a), could be defined as follows:
\begin{equation}
{\small
\begin{split}
&\mathbf{Q} = \text{DwConv}_{3\times3}\big(\text{Conv}_{1\times1}\big(\text{Norm}(\mathbf{prior})\big)\big), \\
&\mathbf{K} = \text{DwConv}_{3\times3}\big(\text{Conv}_{1\times1}\big(\text{Norm}(\mathbf{x})\big)\big), \\
&\mathbf{V} = \text{DwConv}_{3\times3}\big(\text{Conv}_{1\times1}\big(\text{Norm}(\mathbf{x})\big)\big),
\end{split}
}
\label{eq:act_qkv}
\end{equation}
where $\mathbf{prior}$ represents the feature maps of the Color Balance Prior, which will be detailed in the following subsection, and $\mathbf{x}$ denotes the image feature maps output from the previous encoder module. Here, $\text{Norm}$ denotes Layer Normalization, while $\text{Conv}_{1\times1}$ and $\text{DwConv}_{3\times3}$ represent a regular convolution with a $1 \times 1$ kernel and a depth-wise convolution with a $3 \times 3$ kernel, respectively.

The overall working process of the Keep Feature Transformer (KFT) and the Self-Attention Transformer (SAT) is similar to that of the ACT shown in Fig. \ref{fig:ACT}. The main difference lies in the $\mathbf{Q}$, $\mathbf{K}$, and $\mathbf{V}$ components used in their respective attention blocks, which correspond to different feature maps, as illustrated in Fig. \ref{fig:attn} and Algorithm \ref{alg_transformer}.

The KFT is designed to operate on the prior features. Its purpose is to retain those features in the prior feature maps that exhibit greater similarity to the image features. By doing so, the KFT suppresses inappropriate information in the prior, acknowledging that the color balance prior may be an inaccurate representation. Therefore, in the attention block of the KFT, the $\mathbf{Q}$ component corresponds to the image features, while the $\mathbf{K}$ and $\mathbf{V}$ components correspond to the prior features, as shown in Fig. \ref{fig:attn}(b). They are defined as:
\begin{equation}
{\small
\begin{split}
&\mathbf{Q} = \text{DwConv}_{3\times3}\big(\text{Conv}_{1\times1}\big(\text{Norm}(\mathbf{x})\big)\big), \\
&\mathbf{K} = \text{DwConv}_{3\times3}\big(\text{Conv}_{1\times1}\big(\text{Norm}(\mathbf{prior})\big)\big), \\
&\mathbf{V} = \text{DwConv}_{3\times3}\big(\text{Conv}_{1\times1}\big(\text{Norm}(\mathbf{prior})\big)\big).
\end{split}
}
\label{eq:kft_qkv}
\end{equation}
All inputs and operations are consistent with those defined in Eq. (\ref{eq:act_qkv}).

In addition to the cross-attention mechanisms employed by the ACT and KFT, which facilitate interaction between the prior and the image features and guide the learning of contextual-level information, self-attention plays a crucial role that should not be overlooked. To address this, we introduce the Self-Attention Transformer (SAT) block. The SAT is designed to capture the inter-channel dependencies within the image features themselves. Therefore, in the attention block of the SAT, the $\mathbf{Q}$, $\mathbf{K}$, and $\mathbf{V}$ components are all derived from embeddings of the input image features, as shown in Fig. \ref{fig:attn}(c). We substitute the following definitions into Eq. (\ref{eq:inter_c_attn}) to obtain the final calculation of the SAT:
\begin{equation}
{\small
\begin{split}
&\mathbf{Q} = \text{DwConv}_{3\times3}\big(\text{Conv}_{1\times1}\big(\text{Norm}(\mathbf{x})\big)\big), \\
&\mathbf{K} = \text{DwConv}_{3\times3}\big(\text{Conv}_{1\times1}\big(\text{Norm}(\mathbf{x})\big)\big), \\
&\mathbf{V} = \text{DwConv}_{3\times3}\big(\text{Conv}_{1\times1}\big(\text{Norm}(\mathbf{x})\big)\big),
\end{split}
}
\label{eq:kft_qkv}
\end{equation}
where all inputs and operations are consistent with those defined in Eq. (\ref{eq:act_qkv}).

The incorporation of these three types of attention mechanisms - cross-attention in ACT and KFT, and self-attention in SAT - enables the Feature Contextualizer to comprehensively capture and refine contextual information at global scale. The cross-attention mechanisms leverage the contextual cues from the color balance prior to guide the learning of the image features, while the self-attention mechanism captures the intrinsic relationships and dependencies within the image features themselves.

\subsubsection{Scale Harmonizer}
\begin{figure}[ht]
\centering
\includegraphics[width=\columnwidth]{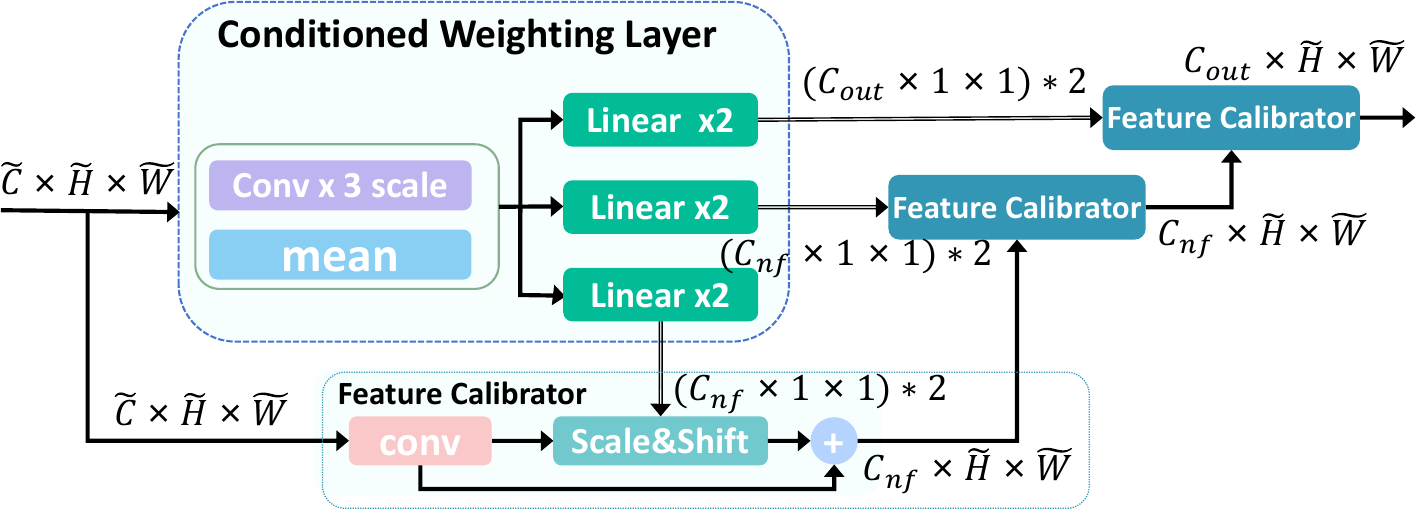}
\caption{Architecture of the Scale Harmonizer module. }
\label{fig:ScaleHarmonizer}
\end{figure}
In the decoding phase of our ``HybSensUIR'' model, the Scale Harmonizers are designed to fuse information from various encoding scales. These are depicted in Fig. \ref{fig:ScaleHarmonizer}.
The features originating from the Feature Contextualizer, or the upsampled, fused features from the lower layer, are initially concatenated with the skip-connected features from the corresponding Detail Restorer at the same scale.
Subsequently, the Scale Harmonizer is employed to further harmonize the concatenated features from various scales.

To harmonize the input feature maps $\mathbf{x} \in \mathbb{R}^{\tilde{C} \times \tilde{H} \times \tilde{W}}$ that originate from the encoder and the lower decoding layer, we designed the Feature Calibrator, which performs feature modulation based on the learnable parameters $\mathbf{P}_{scale}$ and $\mathbf{P}_{shift}$. To ensure that the calibrator is flexible and adaptable to different input features, the scaling and shifting parameters are conditioned by the average of three levels of features extracted at different convolutional granularities from the input feature $\mathbf{x}$. The Feature Calibrator can be depicted as:
\begin{equation}
{\small
\text{Calib}(\mathbf{x}) = Conv2d_{1\times1}(\mathbf{x}) \odot \text{CWL}_{scale}(\mathbf{x}) + \text{CWL}_{shift}(\mathbf{x}),
}
\label{eq:calib}
\end{equation}
where $\mathbf{x}$ is the input feature, $\text{CWL}$ is the Conditioned Weighting Layer that generates the conditioned learnable scale parameter $\mathbf{P}_{scale} \in \mathbb{R}^{C_{nf} \times \tilde{H} \times \tilde{W}}$ and shift parameter $\mathbf{P}{shift} \in \mathbb{R}^{C_{nf} \times \tilde{H} \times \tilde{W}}$. Before $\mathbf{x}$ is calibrated by these two parameters, it first goes through an $1\times1$ convolution to adapt its number of channels to $C_{nf}$ for the subsequent element-wise multiplication and addition.

After sequentially scaling and shifting the features using three groups of Feature Calibrators, we achieve the goal of harmonizing the fused feature channels with learned appropriate weights by adjusting their amplitude and adding biases to optimize their representation.

\subsection{Color Balance Prior}
It is widely recognized that underwater images often exhibit color distortions due to the wavelength-dependent absorption of light, where red light attenuates most rapidly, followed by green and blue~\cite{mobley1993comparison}. This results in greenish or bluish tints. Our objective in correcting this color distortion is to balance the red, green, and blue color channels to achieve a visually harmonious appearance. To facilitate appropriate light compensation, we investigate how illumination affects perceived color in each channel and the relationships between their intensities in air.
Equation~(\ref{eq:colorIntensity}) describes the relationship of the intensity value at pixel position $(x, y)$ in the $i$-th color channel (we consider three channels: R, G, and B) \cite{buchsbaum1980spatial}:
\begin{equation}
{\small
f_i(x, y) = G(x, y)R_i(x, y)I_i(x, y),
}
\label{eq:colorIntensity}
\end{equation}
where $f_i(x, y)$ is the observed pixel intensity, $G(x, y)$ is a factor that depends on the scene geometry, $R_i(x, y)$ is the reflectance of the object at position $(x, y)$ in the $i$-th color channel (representing the true color of this point, excluding illumination), and $I_i(x, y)$ is the illuminant at that point.

In underwater image restoration, the orientation between the subject and the camera is fixed and cannot be adjusted, thus, $G(x, y)$ is a constant factor. Additionally, the reflectance factor $R_i(x, y)$ is inherent and does not change with illumination. Therefore, to calibrate the distribution of color intensity in the UIR process and compensate for illumination attenuation, we focus on estimating and adjusting the $I_i(x, y)$ component.

The average intensity $a_i$ of each channel over an image of size $M \times N$ pixels can be computed as:
\begin{equation}
{\small
a_i = \frac{1}{MN} \sum_{x=0}^{M-1} \sum_{y=0}^{N-1} f_i(x, y),
}
\label{eq:a_i_1}
\end{equation}
\begin{equation}
{\small
a_i = \frac{1}{MN} \sum_{x=0}^{M-1} \sum_{y=0}^{N-1} G(x, y)R_i(x, y)I_i(x, y),
}
\label{eq:a_i_2}
\end{equation}
Under the same condition, the illumination $I_i(x, y)$ can be treated as a constant across the image, so we derive: 
\begin{equation}
{\small
a_i = I_i \frac{1}{MN} \sum_{x=0}^{M-1} \sum_{y=0}^{N-1} G(x, y)R_i(x, y),
}
\label{eq:a_i_3}
\end{equation}
Expressing the averaging as an expectation, we have:
\begin{equation}
{\small
a_i \approx I_i E [GR_i] = I_i E[G] E[R_i],
}
\label{eq:a_i_4}
\end{equation}
where $E [GR_i]$ is the expected value of $G \times R_i$. Since there is no correlation between the geometry (shape) and the reflectance (color) of an object, $G$ and $R_i$ can be considered independent variables, allowing us to split the joint expectation into the product of two independent expectations: $E[G], E[R_i]$. 

Assuming that many different colors are present in the scene and each color is equally likely, the reflectance $R_i$ can be considered a random variable uniformly distributed over the range $[0, 1]$. Therefore, the expected value of the reflectance is:
\begin{equation}
E[R_i] = \int_{0}^{1} r dr = \frac{1}{2}.
\label{eq:reflectance_expectation}
\end{equation}
Substituting back into Equation~(\ref{eq:a_i_4}), we get:
\begin{equation}
{\small
a_i \approx I_i E[G] \left( \frac{1}{2} \right).
}
\label{eq:expect1}
\end{equation}
Assuming a perpendicular orientation between the object and the camera, we have $E[G] = 1$, resulting in:
\begin{equation}
{\small
a_i \approx  \frac{I_i}{2}.
}
\label{eq:expect1}
\end{equation}
This result indicates that the average color intensity $a_i$ of each channel is proportional to its illuminant $I_i$. In an ideal in-air lighting environment, where there is no wavelength-based attenuation, the illuminant of the three channels R, G, and B should be the same, i.e., $I_R = I_G = I_B$. Consequently, the channel-wise average intensities $a_i$ should also be the same across the R, G, and B channels:

\begin{equation}
{\small
a_R \approx a_G \approx a_B,
}
\label{eq:even_intensity}
\end{equation}

where $a_R$, $a_G$, and $a_B$ are the average intensities of the red, green, and blue channels, respectively. This is consistent with the Gray-World Assumption proposed by Buchsbaum~\cite{buchsbaum1980spatial}, which assumes that the average of all color channels in an image is representative of a gray level and estimates the illuminant by computing the global spatial average color.

Therefore, to correct the color cast caused by wavelength-based attenuation in underwater environments, the fundamental approach is to compensate for the attenuated color channels to restore a balanced intensity distribution among the three channels, similar to that in air. Traditional underwater enhancement algorithms demonstrate that this is not a straightforward task. Simple operations often fall short, and even complex methods may struggle with the diverse and challenging underwater environment.

Instead of relying on intricate color balancing techniques, we opt for a simpler approach. We directly use the average of the red, green, and blue channels to establish a basic color balance prior that conforms to the even intensity condition in Equation~(\ref{eq:even_intensity}), as defined in Equation~(\ref{eq:prior}). This prior serves as a directional guide for our deep learning methods. By introducing the Color Balance Prior, our previously described ``HybSensHIR'' system is transformed into the ``Guided Hybrid Sense of UIR Framework''.
\begin{equation}
{\small
\text{Prior}_i(x,y) = \frac{R(x,y) + G(x,y) + B(x,y)}{3},
}
\label{eq:prior}
\end{equation}
where $\text{Prior}_i(x,y)$ represents the value for the $i$-th channel at pixel $(x,y)$, and all three channels share the same value; $R(x,y)$, $G(x,y)$, and $B(x,y)$ are the red, green, and blue channel values at pixel $(x,y)$, respectively.

Fig. \ref{fig:scatter} illustrates that the orientation of our color balance prior corresponds with the tilt seen in the color distribution comparison between the original and the restored underwater images. This prior's position at the heart of the restored image's scatter plot is crucial. Its central location signals alignment with the mean of our intended restoration outcomes, underscoring its potential to effectively steer the restoration process. Consequently, we expect this color balance prior to lead our model towards enhanced and consistent results.

We do not use the color balance prior value directly; instead, we employ a Feature Extractor to extract feature embeddings before utilizing them, as shown in Fig. \ref{fig:overall}. This approach is taken because directly using the prior value may not capture the necessary contextual relationships and intricate patterns required for accurate color correction. Due to the excellent feature extraction capabilities of the Nonlinear Activation-Free Block (NAFB) used in the Detail Restorer, we employ the NAFB to serve as the Feature Extractor.

We integrate this color balance prior at two critical points to guide the color improvement trajectory: within the Feature Contextualizer, as depicted in the previous subsection, and at the topmost Feature Harmonizer. Given that the Feature Contextualizer focuses on long-range relationships to correct global color features, we employ the color balance prior as a robust guide through in-depth interaction with the image features using two types of cross-attention mechanisms, as illustrated in Fig. \ref{fig:FeatureContextualizer} to Fig. \ref{fig:ACT}. 

However, considering the approximate nature and lack of precision of the color balance prior, it is merged only once with the features from the Detail Restorer and the preceding Feature Harmonizer, just before the final Feature Harmonizer. This serves as a subtle reminder of the color trends, ensuring that the final output maintains a balance between the learned features and the color prior.

\begin{table*}[ht]
\caption{Quantitative comparison on the paired test sets UIEB, EUVP and LSUI based on referenced metrics. The table is organized with traditional methods listed in the upper section and deep learning methods detailed in the lower section. Due to the large margin between the metrics of the deep learning methods and the traditional methods, the highest-performing results and the second-best results are highlighted in bold and underlined, respectively, within each section for better comparison.}
\centering
\adjustbox{width=0.85\textwidth}{%
\begin{tabular}{lrrrrrrrrrrr}
\toprule
\multirow{2}{*}{\textbf{Method}} & \multicolumn{2}{c}{\textbf{Computational Cost}} & \multicolumn{3}{c}{\textbf{UIEB}}                         & \multicolumn{3}{c}{\textbf{EUVP}}                        & \multicolumn{3}{c}{\textbf{LSUI}}                        \\ \cmidrule(lr){2-3} \cmidrule(lr){4-6} \cmidrule(lr){7-9}  \cmidrule(lr){10-12}
                        & MACs(G)$\downarrow$ & Params(M)$\downarrow$ & PSNR$\uparrow$ & SSIM$\uparrow$ & LPIPS$\downarrow$ & PSNR$\uparrow$ & SSIM$\uparrow$ & LPIPS$\downarrow$ & PSNR$\uparrow$ & SSIM$\uparrow$ & LPIPS$\downarrow$ \\ \midrule
\textbf{WCID} \cite{chiang2011underwater}                    & \multicolumn{1}{c}{-}                      & \multicolumn{1}{c}{-}                  & 11.65          & 0.322          & 0.424           & 12.93          & 0.270          & 0.480           & 12.38          & 0.285          & 0.444           \\
\textbf{Fusion} \cite{ancuti2012enhancing}                  & \multicolumn{1}{c}{-}                      & \multicolumn{1}{c}{-}                  & 15.00          & 0.726          & 0.373           & 16.07          & 0.703          & 0.372           & 16.09          & 0.756          & 0.324           \\
\textbf{GBD\&RC} \cite{li2016single}                 & \multicolumn{1}{c}{-}                      & \multicolumn{1}{c}{-}                  & 12.35          & 0.625          & 0.446           & 14.20          & 0.598          & 0.449           & 13.59          & 0.648          & 0.460           \\
\textbf{min\_info\_loss} \cite{li2016underwater}          & \multicolumn{1}{c}{-}                      & \multicolumn{1}{c}{-}                  & 17.13    & 0.783          & 0.327           & 15.23          & 0.643          & 0.439           & 16.41          & 0.720          & 0.382           \\
\textbf{IBLA} \cite{peng2017underwater}                    & \multicolumn{1}{c}{-}                      & \multicolumn{1}{c}{-}                  & 15.93          & 0.710          & 0.291           & \textbf{18.91} & \underline{0.710}    & \textbf{0.319}  & 16.95          & 0.721          & 0.331           \\
\textbf{Sea-thru} \cite{akkaynak2019sea}                 & \multicolumn{1}{c}{-}                      & \multicolumn{1}{c}{-}                  & 13.82          & 0.580          & 0.421           & 12.72          & 0.499          & 0.496           & 12.91          & 0.505          & 0.501           \\
\textbf{UNTV} \cite{xie2021variational}                    & \multicolumn{1}{c}{-}             & \multicolumn{1}{c}{-}               & 16.46 & 0.669          & 0.420           & \underline{17.63}    & 0.611          & 0.335           & \textbf{18.36} & 0.660          & 0.376           \\
\textbf{HLRP} \cite{zhuang2022underwater}                    & \multicolumn{1}{c}{-}                      & \multicolumn{1}{c}{-}                  & 13.30          & 0.259          & 0.364           & 11.41          & 0.186          & 0.500           & 12.96          & 0.221          & 0.429           \\
\textbf{MLLE} \cite{zhang2022underwater}                    & \multicolumn{1}{c}{-}                      & \multicolumn{1}{c}{-}                  & \textbf{18.74} & 0.814          & 0.234           & 15.14          & 0.633          & \underline{0.323}     & 17.87          & 0.730          & \textbf{0.278}  \\
\textbf{ROP} \cite{liu2022rank}                     & \multicolumn{1}{c}{-}                      & \multicolumn{1}{c}{-}                  & 18.48          & \textbf{0.849} & \textbf{0.209}  & 15.34          & \textbf{0.714} & 0.343           & 17.38          & \textbf{0.806} & \underline{0.281}     \\
\textbf{ROP+} \cite{liu2022rank}                    & \multicolumn{1}{c}{-}                      & \multicolumn{1}{c}{-}                  & 15.88          & 0.776          & 0.287           & 13.46          & 0.613          & 0.393           & 14.51          & 0.692          & 0.354           \\
\textbf{ADPCC} \cite{zhou2023underwater}                   & \multicolumn{1}{c}{-}                      & \multicolumn{1}{c}{-}                  & 17.33          & 0.819          & 0.219           & 15.20          & 0.692          & 0.349           & 16.20          & \underline{0.763}    & 0.299           \\
\textbf{ICSP} \cite{hou2023non}                    & \multicolumn{1}{c}{-}                      & \multicolumn{1}{c}{-}                  & 12.04          & 0.599          & 0.552           & 11.73          & 0.522          & 0.413           & 11.96          & 0.583          & 0.508           \\
\textbf{WWPF} \cite{zhang2023underwater}                    & \multicolumn{1}{c}{-}                      & \multicolumn{1}{c}{-}                  & \underline{18.60}    & \underline{0.822}    & \underline{0.218}     & 15.95          & 0.648          & 0.337           & \underline{17.90}    & 0.739          & 0.283           \\ \midrule
\textbf{UGAN} \cite{fabbri2018enhancing}                    & 19.82                 & 57.17                   & 21.52          & 0.804          & 0.189           & 23.30          & 0.815          & 0.220           & 24.22          & 0.840          & 0.191           \\
\textbf{WaterNet} \cite{li2019underwater}                & 71.42                 & 1.09                    & 22.82          & 0.907          & 0.125           & 24.12          & 0.839          & 0.222           & 25.25          & 0.877          & 0.164           \\
\textbf{UWCNN} \cite{li2020underwater}                   & \textbf{2.61}         & \textbf{0.04}           & 18.44          & 0.844          & 0.203           & 23.49          & 0.830          & 0.231           & 21.73          & 0.844          & 0.228           \\
\textbf{Shallow-UWNet} \cite{naik2021shallow}           & 21.63                 & \underline{0.22}              & 18.30          & 0.846          & 0.206           & 23.59          & 0.832          & 0.228           & 21.91          & 0.846          & 0.233           \\
\textbf{Ucolor} \cite{li2021underwater}                  & 1002.00               & 105.51                  & 18.37          & 0.814          & 0.221           & 23.72          & 0.828          & 0.205           & 21.30          & 0.821          & 0.225           \\
\textbf{CLUIE-Net} \cite{li2022beyond}                & 31.13                                 & 13.40                                   & 19.95          & 0.874          & 0.168           & 24.85          & 0.844          & 0.186           & 23.57          & 0.864          & 0.175           \\
\textbf{PUIE-Net} \cite{fu2022uncertainty}                 & 30.09                                 & 1.40                                    & 21.04          & 0.877          & 0.136           & 20.48          & 0.784          & 0.270           & 22.07          & 0.864          & 0.191           \\
\textbf{STSC} \cite{wang2022semantic}                     & 204.84                                & 32.92                                   & 21.20          & 0.820          & 0.183           & 25.08          & 0.844          & 0.176           & 24.26          & 0.851          & 0.181           \\
\textbf{TACL} \cite{liu2022twin}                     & 120.03                                & 28.29                                   & 19.83          & 0.761          & 0.222           & 20.99          & 0.782          & 0.213           & 22.97          & 0.828          & 0.176           \\
\textbf{UIE-WD} \cite{ma2022wavelet}                   & 51.38                                 & 14.49                                   & 20.28          & 0.848          & 0.198           & 17.80          & 0.760          & 0.292           & 19.23          & 0.803          & 0.284           \\
\textbf{URSCT} \cite{ren2022reinforced}                   & 18.11                                 & 11.26                                   & 22.77          & 0.915          & 0.120           & \underline{25.74}    & \textbf{0.855} & 0.180           & \underline{25.87}    & \textbf{0.883} & \underline{0.146}     \\
\textbf{USUIR} \cite{fu2022unsupervised}                   & 14.81                                 & 0.23                                    & 22.48          & 0.907          & 0.124           & 21.94    & 0.810 & 0.239           & 23.75    & 0.860 & 0.184     \\
\textbf{CCMSRNet} \cite{qi2023deep}                 & 43.60                                 & 21.13                                   & 17.04          & 0.790          & 0.322           & 17.77          & 0.720          & 0.381           & 18.54          & 0.788          & 0.352           \\
\textbf{DeepWaveNet} \cite{sharma2023wavelength}              & 18.15                                 & 0.28                                    & 21.55 & 0.904          & 0.143           & 23.41          & 0.836          & 0.204     & 23.81          & 0.870          & 0.177           \\
\textbf{GUPDM} \cite{mu2023generalized}                   & 95.80                                 & 1.49                                    & 22.13          & 0.903    & 0.131     & 24.79          & 0.847    & 0.184           & 25.33          & 0.877          & 0.150           \\
\textbf{MBANet} \cite{xue2023investigating}                  & 11.30                                 & 0.52                                    & 19.58          & 0.800          & 0.209           & 23.76          & 0.819    & 0.225           & 23.02          & 0.843          & 0.210           \\
\textbf{$ \mathbf{NU^2Net}$ } \cite{guo2023underwater}                  & 10.49                                 & 3.15                                    & 19.99          & 0.850          & 0.196           & 21.51          & 0.810          & 0.308           & 22.13          & 0.872          & 0.243           \\
\textbf{PUGAN} \cite{cong2023pugan}                    & 75.40                                 & 101.19                                  & 20.52          & 0.812          & 0.216           & 22.58          & 0.820          & 0.212           & 23.14          & 0.836          & 0.216           \\
\textbf{Semi-UIR} \cite{huang2023contrastive}                 & 72.88                                 & 3.31                                    & \textbf{23.64} & 0.888          & 0.120           & 24.59          & 0.821          & \underline{0.172}     & 25.40          & 0.843          & 0.160           \\
\textbf{SyreaNet} \cite{wen2023syreanet}                & 140.88                                & 29.05                                   & 22.72          & \underline{0.918}    & \underline{0.116}     & 23.25          & 0.834          & 0.224           & 24.90          & 0.872          & 0.168           \\
\textbf{TUDA} \cite{wang2023domain}                     & 85.43                           & 2.73                                    & 22.72          & 0.915          & 0.118           & 23.73          & 0.843          & 0.207           & 25.52          & 0.878          & 0.154           \\
\textbf{U-Transformer} \cite{peng2023u}           & \underline{2.98}                            & 22.82                                   & 20.75          & 0.810          & 0.228           & 24.99          & 0.829          & 0.238           & 25.15          & 0.838          & 0.221           \\
\textbf{SFGNet} \cite{zhao2024toward}                  & 81.58                                 & 1.30                                    & 19.57          & 0.685          & 0.214           & 22.68          & 0.585          & 0.221           & 22.71          & 0.653          & 0.204           \\
\textbf{Ours}                     & 10.05                 & 1.15                    & \underline{23.63}    & \textbf{0.923} & \textbf{0.100}          & \textbf{25.75} & \underline{0.847}    & \textbf{0.111}     & \textbf{26.39} & \underline{0.880}    & \textbf{0.098}     \\
\bottomrule
\end{tabular}%
}
\label{tab:quan1}
\end{table*}

\begin{table}[ht]
\caption{Quantitative comparison on the paired test sets UIEB, EUVP and LSUI based on non-reference metrics.}
\adjustbox{width=\columnwidth}{%
\begin{tabular}{lcccccc}
\toprule
\multirow{2}{*}{\textbf{Method}}  & \multicolumn{2}{c}{\textbf{UIEB}}                         & \multicolumn{2}{c}{\textbf{EUVP}}                        & \multicolumn{2}{c}{\textbf{LSUI}}                        \\ 
\cmidrule(lr){2-3} \cmidrule(lr){4-5} \cmidrule(lr){6-7}
& UCIQE$\uparrow$ & UIQM$\uparrow$ & UCIQE$\uparrow$ & UIQM$\uparrow$  & UCIQE$\uparrow$ & UIQM$\uparrow$ \\ 
\midrule
\textbf{WCID} \cite{chiang2011underwater}   &0.453          & 2.66          & 0.491          & 2.64          & 0.450          & 2.81          \\
\textbf{Fusion} \cite{ancuti2012enhancing}  &0.342          & 2.75          & 0.411          & 2.75          & 0.394          & 3.17          \\
\textbf{GBD\&RC} \cite{li2016single}    &\underline{0.509}    & 3.04          & \textbf{0.575} & 2.53          & \textbf{0.547} & 3.15          \\
\textbf{min\_info\_loss} \cite{li2016underwater}    &\textbf{0.544} & \textbf{3.63} & 0.497          & 2.96          & 0.498          & 3.25          \\
\textbf{IBLA} \cite{peng2017underwater} &0.447          & 2.30          & 0.483          & 2.19          & 0.461          & 2.60          \\
\textbf{Sea-thru} \cite{akkaynak2019sea}    &0.502          & 3.11          & \underline{0.518}    & 2.46          & \underline{0.518}    & 2.89          \\
\textbf{UNTV} \cite{xie2021variational} &0.436          & 3.09          & 0.481          & 2.46          & 0.449          & 3.22          \\
\textbf{HLRP} \cite{zhuang2022underwater}   &0.476          & 2.66          & 0.509          & 2.87          & 0.503          & 2.89          \\
\textbf{MLLE} \cite{zhang2022underwater}    &0.444          & 3.04          & 0.474          & 2.30          & 0.457          & 3.09          \\
\textbf{ROP} \cite{liu2022rank} &0.454          & \underline{3.51}    & 0.441          & \textbf{3.15} & 0.448          & \textbf{3.47} \\
\textbf{ROP+} \cite{liu2022rank}    &0.449          & 3.38          & 0.450          & \underline{3.06}    & 0.443          & \underline{3.40}    \\
\textbf{ADPCC} \cite{zhou2023underwater}    &0.485          & 2.84          & 0.495          & 2.29          & 0.481          & 2.77          \\
\textbf{ICSP} \cite{hou2023non} &0.408          & 2.76          & 0.443          & 2.09          & 0.399          & 2.99          \\
\textbf{WWPF} \cite{zhang2023underwater}    &0.446          & 3.15          & 0.467          & 2.62    & 0.456          & 3.18          \\
\midrule
\textbf{UGAN} \cite{fabbri2018enhancing}    &0.427          & 3.38          & \textbf{0.442} & 3.00          & \textbf{0.433} & 3.14          \\
\textbf{WaterNet} \cite{li2019underwater}   &\underline{0.436}    & 3.26          & 0.426          & 2.99          & 0.418          & 3.08          \\
\textbf{UWCNN} \cite{li2020underwater}  &0.367          & 3.13          & 0.407          & 3.04          & 0.360          & 3.11          \\
\textbf{Shallow-UWNet} \cite{naik2021shallow}   &0.355          & 3.19          & 0.401          & 3.05          & 0.352          & 3.12          \\
\textbf{Ucolor} \cite{li2021underwater} &0.370          & 3.04          & 0.408          & 3.05          & 0.362          & 3.06          \\
\textbf{CLUIE-Net} \cite{li2022beyond}  &0.403          & 3.16          & 0.415          & 2.98          & 0.396          & 3.07          \\
\textbf{PUIE-Net} \cite{fu2022uncertainty}  &0.370          & \underline{3.54}    & 0.391          & 3.20          & 0.378          & \underline{3.53}    \\
\textbf{STSC} \cite{wang2022semantic}   &0.415          & 3.11          & 0.414          & 2.94          & 0.402          & 3.07          \\
\textbf{TACL} \cite{liu2022twin}    &0.427          & 3.23          & 0.440          & \underline{3.27}    & \textbf{0.433} & 3.33          \\
\textbf{UIE-WD} \cite{ma2022wavelet}    &0.407          & 3.16          & 0.440          & 2.92          & 0.410          & 3.03          \\
\textbf{URSCT} \cite{ren2022reinforced} &0.432          & 3.19          & 0.425          & 2.90          & 0.417          & 3.05          \\
\textbf{USUIR} \cite{fu2022unsupervised}    &0.427          & 3.23          & 0.408          & 3.09          & 0.412          & 3.12          \\
\textbf{CCMSRNet} \cite{qi2023deep} &0.424          & \textbf{3.73} & 0.436          & \textbf{3.38} & 0.430          & \textbf{3.57} \\
\textbf{DeepWaveNet} \cite{sharma2023wavelength}    &0.424          & 3.13          & 0.426          & 3.01          & 0.415          & 3.06          \\
\textbf{GUPDM} \cite{mu2023generalized} &0.427          & 3.14          & 0.431          & 2.94          & 0.420          & 3.06          \\
\textbf{MBANet} \cite{xue2023investigating} &0.414          & 3.39          & 0.436          & 3.26          & 0.420          & 3.35          \\
\textbf{NU\textsuperscript{2}Net} \cite{guo2023underwater}  &0.390          & 3.52          & 0.419          & 3.12          & 0.396          & 3.46          \\
\textbf{PUGAN} \cite{cong2023pugan}     &0.418          & 3.29          & 0.425          & 3.04          & 0.413          & 3.15          \\
\textbf{Semi-UIR} \cite{huang2023contrastive}   &0.428          & 3.17          & 0.424          & 2.94          & 0.419          & 3.05          \\
\textbf{SyreaNet} \cite{wen2023syreanet}    &\textbf{0.442} & 3.12          & 0.418          & 2.99          & 0.422          & 3.08          \\
\textbf{TUDA} \cite{wang2023domain} &0.429          & 3.14          & 0.415          & 2.94          & 0.417          & 3.05          \\
\textbf{U-Transformer} \cite{peng2023u} &0.434          & 3.11          & 0.434          & 2.92          & 0.427          & 3.05          \\
\textbf{SFGNet} \cite{zhao2024toward}   &0.432          & 2.99          & \textbf{0.442} & 2.84          & 0.426          & 3.03          \\
\textbf{Ours}   &0.433          & 3.22          & \underline{0.441}    & 2.90          & \underline{0.432}    & 3.06          \\
\bottomrule
\end{tabular}%
}
\label{tab:quan2}
\end{table}

\subsection{Loss Function}
In our methodology, we aim to restore the visual quality of underwater images while preserving their important features. To achieve this, we have developed a composite loss function that combines pixel-level fidelity loss $\mathcal{L}_{f}$, structural similarity loss $\mathcal{L}_{s}$, and perceptual quality loss $\mathcal{L}_{p}$, formulated as:
\begin{equation}
{\small
\mathcal{L} = w_1 \cdot \mathcal{L}_{f} + w_2 \cdot \mathcal{L}_{s} + w_3 \cdot \mathcal{L}_{p}.
}
\label{eq:3}
\end{equation}
We empirically set $w_1 = 1$, $w_2 = 0.3$, and $w_3 = 0.7$.

$\mathcal{L}_{f}$ is dedicated to ensuring high fidelity at the pixel level. To quantify pixel-level discrepancies between our model's output and the reference image, we employ the SmoothL1Loss function, balancing sensitivity to large errors with robustness against outliers:
\begin{equation}
{\small
\mathcal{L}_{f}(R, T) = \frac{1}{N} \sum_{i=1}^N \begin{cases}
\frac{1}{2\beta} (R_i - T_i)^2, & \text{if } |R_i - T_i| < \beta \\
|R_i - T_i| - \frac{\beta}{2}, & \text{otherwise}.
\end{cases}
}
\label{eq:Lf}
\end{equation}
In this equation, $R$ represents the restored image, and $T$ denotes the target reference image. The variable $N$ is the total number of pixel elements in the images. The parameter $\beta$ is a positive threshold that controls the transition point where the loss function changes from quadratic to linear behavior; we use the common default value of $\beta = 1.0$.

To maintain similarity in terms of luminance, contrast, and structure, we define $\mathcal{L}_{s}$ as:
\begin{equation}
{\small
\mathcal{L}_{s}(R, T) = 1 - \frac{(2 \mu_R \mu_T + C_1)(2 \sigma_{RT} + C_2)}{(\mu_R^2 + \mu_T^2 + C_1)(\sigma_R^2 + \sigma_T^2 + C_2)},
}
\label{eq:4}
\end{equation}
where $\mu_R$ and $\mu_T$ represent the average pixel values of the output restored image $R$ and the target reference image $T$, $\sigma_R^2$ and $\sigma_T^2$ are the variances of images $R$ and $T$, ad $\sigma_{RT}$ is the covariance between $R$ and $T$. $C_1$ and $C_2$ are constants to stabilize the division with weak denominators. The second part of this equation corresponds to Structural Similarity Index Measure \cite{wang2004image}.

To ensure that the restored images align with human visual perception, we incorporate the Learned Perceptual Image Patch Similarity (LPIPS)~\cite{zhang2018unreasonable} into the loss function $\mathcal{L}_{p}$, defined as:
\begin{equation}
{\small
\mathcal{L}_{p}(R, T) = \mathrm{LPIPS}(R, T).
}
\label{eq:L_p}
\end{equation}
In this equation, $\mathrm{LPIPS}(R, T)$ computes the perceptual distance between the restored image $R$ and the target reference image $T$ using features extracted from a pretrained AlexNet model on the ImageNet dataset.

\section{Experiments}
\begin{figure*}[ht]
    \begin{minipage}[b]{1.0\linewidth}
        \begin{minipage}[b]{0.119\linewidth}
            \centering
            \centerline{\includegraphics[width=\linewidth]{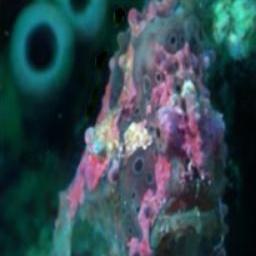}}
            \vspace{-2mm}
            \centerline{{\scriptsize UCIQE / UIQM}}
        \end{minipage}   
        \hfill
        \begin{minipage}[b]{0.119\linewidth}
            \centering
            \centerline{\includegraphics[width=\linewidth]{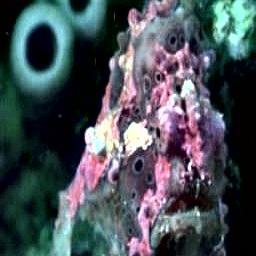}}
            \vspace{-2mm}
            \centerline{{\scriptsize 0.509 / 2.13}}
        \end{minipage}
        \hfill
        \begin{minipage}[b]{0.119\linewidth}
            \centering
            \centerline{\includegraphics[width=\linewidth]{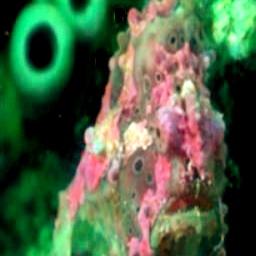}}
            \vspace{-2mm}
            \centerline{{\scriptsize \textbf{0.548} / 2.27}}
        \end{minipage}   
        \hfill
        \begin{minipage}[b]{0.119\linewidth}
            \centering
            \centerline{\includegraphics[width=\linewidth]{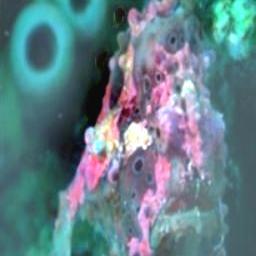}}
            \vspace{-2mm}
            \centerline{{\scriptsize 0.424 / \textbf{3.19}}}
        \end{minipage}  
        \hfill
        \begin{minipage}[b]{0.119\linewidth}
            \centering
            \centerline{\includegraphics[width=\linewidth]{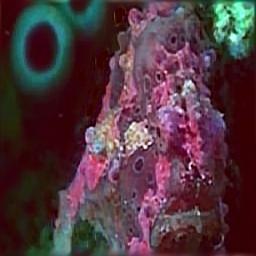}}
            \vspace{-2mm}
            \centerline{{\scriptsize 0.537 / 2.37}}
        \end{minipage}
        \hfill
        \begin{minipage}[b]{0.119\linewidth}
            \centering
            \centerline{\includegraphics[width=\linewidth]{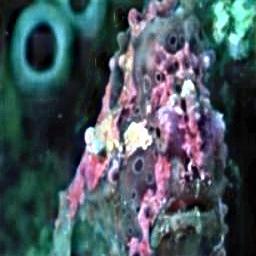}}
            \vspace{-2mm}
            \centerline{{\scriptsize 0.464 / 2.84}}
        \end{minipage}
        \hfill
        \begin{minipage}[b]{0.119\linewidth}
            \centering
            \centerline{\includegraphics[width=\linewidth]{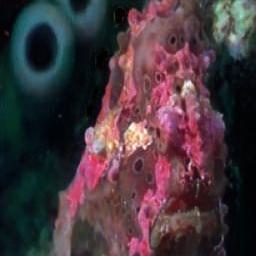}}
            \vspace{-2mm}
            \centerline{{\scriptsize 0.522 / 2.72}}
        \end{minipage}
        \hfill
        \begin{minipage}[b]{0.119\linewidth}
            \centering
            \centerline{\includegraphics[width=\linewidth]{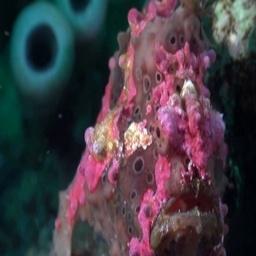}}
            \vspace{-2mm}
            \centerline{{\scriptsize 0.502 / 2.88}}
        \end{minipage}
    \end{minipage}
    \begin{minipage}[b]{1.0\linewidth}
        \begin{minipage}[b]{0.119\linewidth}
            \centering
            \centerline{\includegraphics[width=\linewidth]{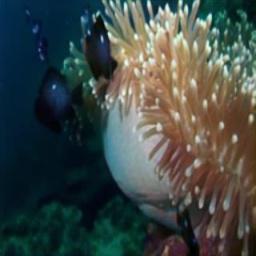}}
            \vspace{-2mm}
            \centerline{{\scriptsize UCIQE / UIQM}}
        \end{minipage}   
        \hfill
        \begin{minipage}[b]{0.119\linewidth}
            \centering
            \centerline{\includegraphics[width=\linewidth]{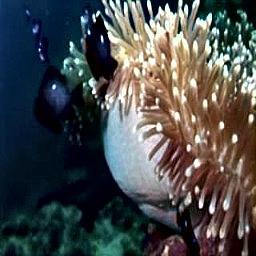}}
            \vspace{-2mm}
            \centerline{{\scriptsize 0.550 / 2.10}}
        \end{minipage}
        \hfill
        \begin{minipage}[b]{0.119\linewidth}
            \centering
            \centerline{\includegraphics[width=\linewidth]{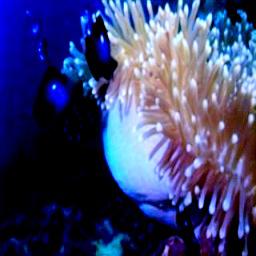}}
            \vspace{-2mm}
            \centerline{{\scriptsize 0.565 / 1.63}}
        \end{minipage}   
        \hfill
        \begin{minipage}[b]{0.119\linewidth}
            \centering
            \centerline{\includegraphics[width=\linewidth]{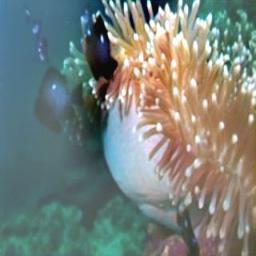}}
            \vspace{-2mm}
            \centerline{{\scriptsize 0.442 / \textbf{3.12}}}
        \end{minipage}  
        \hfill
        \begin{minipage}[b]{0.119\linewidth}
            \centering
            \centerline{\includegraphics[width=\linewidth]{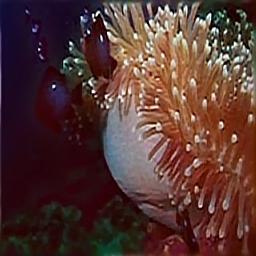}}
            \vspace{-2mm}
            \centerline{{\scriptsize \textbf{0.592} / 2.07}}
        \end{minipage}
        \hfill
        \begin{minipage}[b]{0.119\linewidth}
            \centering
            \centerline{\includegraphics[width=\linewidth]{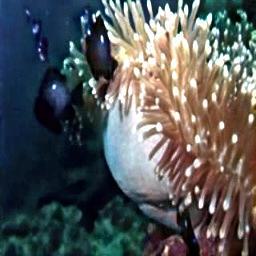}}
            \vspace{-2mm}
            \centerline{{\scriptsize 0.517 / 2.71}}
        \end{minipage}
        \hfill
        \begin{minipage}[b]{0.119\linewidth}
            \centering
            \centerline{\includegraphics[width=\linewidth]{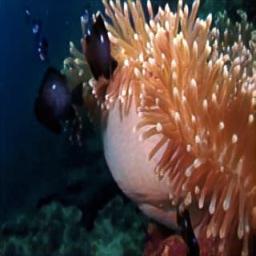}}
            \vspace{-2mm}
            \centerline{{\scriptsize 0.559 / 2.51}}
        \end{minipage}
        \hfill
        \begin{minipage}[b]{0.119\linewidth}
            \centering
            \centerline{\includegraphics[width=\linewidth]{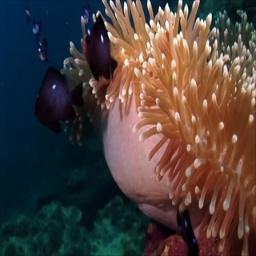}}
            \vspace{-2mm}
            \centerline{{\scriptsize 0.567 / 2.58}}
        \end{minipage}
    \end{minipage}    
    \begin{minipage}[b]{1.0\linewidth}
        \begin{minipage}[b]{0.119\linewidth}
            \centering
            \centerline{\includegraphics[width=\linewidth]{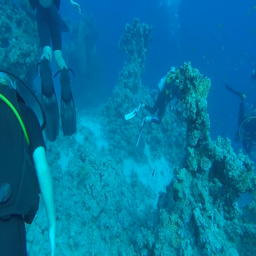}}
            \vspace{-2mm}
            \centerline{{\scriptsize UCIQE / UIQM}}
        \end{minipage}   
        \hfill
        \begin{minipage}[b]{0.119\linewidth}
            \centering
            \centerline{\includegraphics[width=\linewidth]{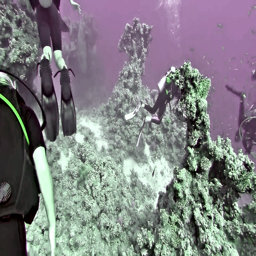}}
            \vspace{-2mm}
            \centerline{{\scriptsize 0.432 / 2.88}}
        \end{minipage}
        \hfill
        \begin{minipage}[b]{0.119\linewidth}
            \centering
            \centerline{\includegraphics[width=\linewidth]{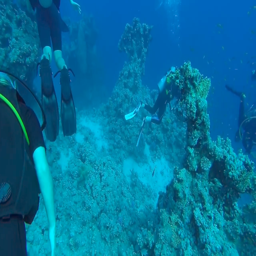}}
            \vspace{-2mm}
            \centerline{{\scriptsize 0.452 / 1.35}}
        \end{minipage}   
        \hfill
        \begin{minipage}[b]{0.119\linewidth}
            \centering
            \centerline{\includegraphics[width=\linewidth]{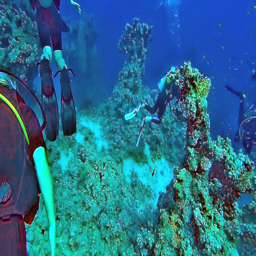}}
            \vspace{-2mm}
            \centerline{{\scriptsize \textbf{0.493} / \textbf{3.32}}}
        \end{minipage}  
        \hfill
        \begin{minipage}[b]{0.119\linewidth}
            \centering
            \centerline{\includegraphics[width=\linewidth]{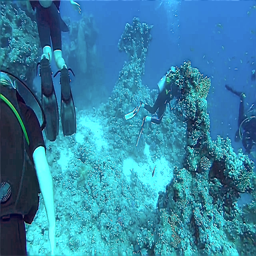}}
            \vspace{-2mm}
            \centerline{{\scriptsize 0.438 / 2.17}}
        \end{minipage}
        \hfill
        \begin{minipage}[b]{0.119\linewidth}
            \centering
            \centerline{\includegraphics[width=\linewidth]{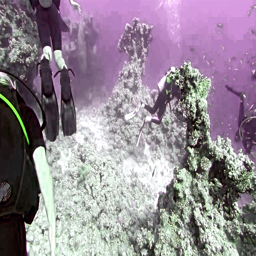}}
            \vspace{-2mm}
            \centerline{{\scriptsize 0.453 / 3.09}}
        \end{minipage}
        \hfill
        \begin{minipage}[b]{0.119\linewidth}
            \centering
            \centerline{\includegraphics[width=\linewidth]{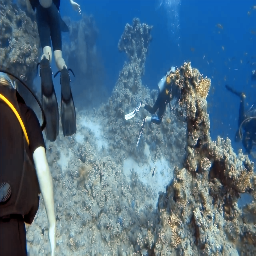}}
            \vspace{-2mm}
            \centerline{{\scriptsize 0.443 / 2.95}}
        \end{minipage}
        \hfill
        \begin{minipage}[b]{0.119\linewidth}
            \centering
            \centerline{\includegraphics[width=\linewidth]{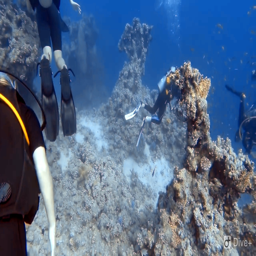}}
            \vspace{-2mm}
            \centerline{{\scriptsize 0.451 / 3.05}}
        \end{minipage}
    \end{minipage}

    \begin{minipage}[b]{1.0\linewidth}
        \begin{minipage}[b]{0.119\linewidth}
            \centering
            \centerline{\includegraphics[width=\linewidth]{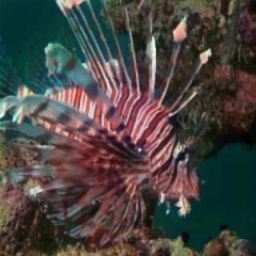}}
            \vspace{-2mm}
            \centerline{{\scriptsize UCIQE / UIQM}}
            \centerline{{\footnotesize (a)Input}}\medskip
        \end{minipage}   
        \hfill
        \begin{minipage}[b]{0.119\linewidth}
            \centering
            \centerline{\includegraphics[width=\linewidth]{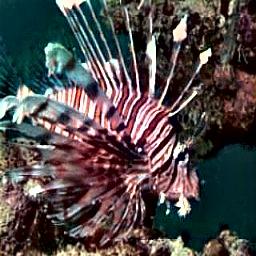}}
            \vspace{-2mm}
            \centerline{{\scriptsize 0.577 / 1.54}}
            \centerline{{\footnotesize (b)MLLE}}\medskip
        \end{minipage}
        \hfill
        \begin{minipage}[b]{0.119\linewidth}
            \centering
            \centerline{\includegraphics[width=\linewidth]{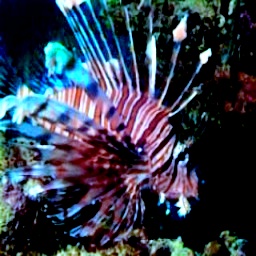}}
            \vspace{-2mm}
            \centerline{{\scriptsize 0.579 / 0.87}}
            \centerline{{\footnotesize (c)IBLA}}\medskip
        \end{minipage}   
        \hfill
        \begin{minipage}[b]{0.119\linewidth}
            \centering
            \centerline{\includegraphics[width=\linewidth]{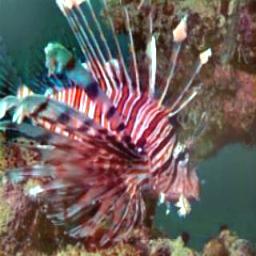}}
            \vspace{-2mm}
            \centerline{{\scriptsize 0.497 / \textbf{3.32}}}
            \centerline{{\footnotesize (d)ROP}}\medskip
        \end{minipage}  
        \hfill
        \begin{minipage}[b]{0.119\linewidth}
            \centering
            \centerline{\includegraphics[width=\linewidth]{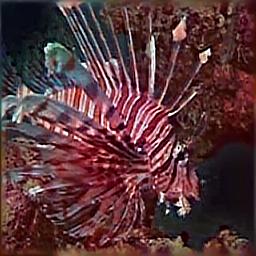}}
            \vspace{-2mm}
            \centerline{{\scriptsize \textbf{0.584} / 1.67}}
            \centerline{{\footnotesize (e)UNTV}}\medskip
        \end{minipage}
        \hfill
        \begin{minipage}[b]{0.119\linewidth}
            \centering
            \centerline{\includegraphics[width=\linewidth]{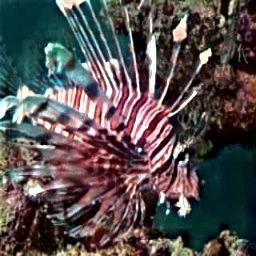}}
            \vspace{-2mm}
            \centerline{{\scriptsize 0.567 / 1.98}}
            \centerline{{\footnotesize (f)WWPF}}\medskip
        \end{minipage}
        \hfill
        \begin{minipage}[b]{0.119\linewidth}
            \centering
            \centerline{\includegraphics[width=\linewidth]{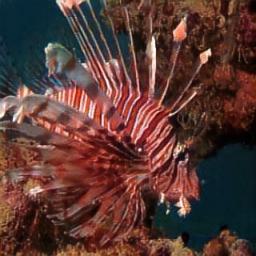}}
            \vspace{-2mm}
            \centerline{{\scriptsize 0.557 / 2.85}}
            \centerline{{\footnotesize (g)Ours}}\medskip
        \end{minipage}
        \hfill
        \begin{minipage}[b]{0.119\linewidth}
            \centering
            \centerline{\includegraphics[width=\linewidth]{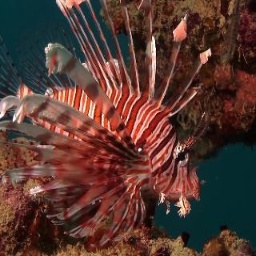}}
            \vspace{-2mm}
            \centerline{{\scriptsize 0.538 / 2.92}}
            \centerline{{\footnotesize (g)Reference}}\medskip
        \end{minipage}
    \end{minipage}
    \caption{Visual comparison with traditional UIR methods. The results in columns (b) to (f) reveal inadequate color correction, often retaining or introducing color distortions and over-exposure. In contrast, our method effectively corrects colors and restores details, significantly outperforming traditional approaches in both color accuracy and texture preservation. }
    
    \label{fig:traditional-compare}
\end{figure*}
\begin{figure*}[ht]
    \begin{minipage}[b]{1.0\linewidth}
        \begin{minipage}[b]{0.119\linewidth}
            \centering
            \centerline{\includegraphics[width=\linewidth]{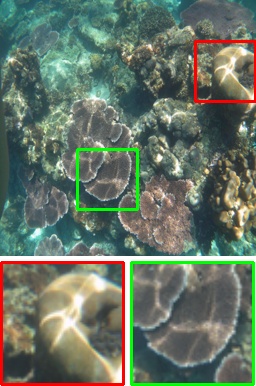}}
        \end{minipage}   
        \hfill
        \begin{minipage}[b]{0.119\linewidth}
            \centering
            \centerline{\includegraphics[width=\linewidth]{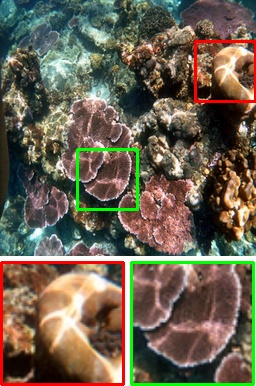}}
        \end{minipage}
        \hfill
        \begin{minipage}[b]{0.119\linewidth}
            \centering
            \centerline{\includegraphics[width=\linewidth]{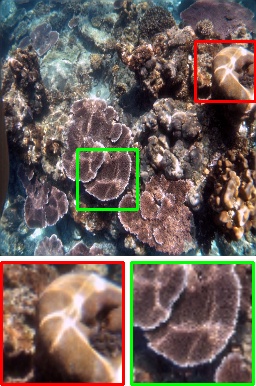}}
        \end{minipage}
        \hfill
        \begin{minipage}[b]{0.119\linewidth}
            \centering
            \centerline{\includegraphics[width=\linewidth]{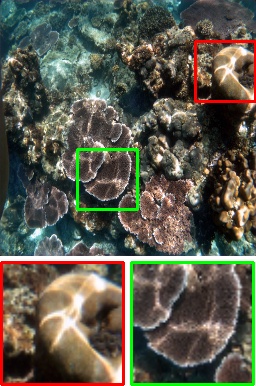}}
        \end{minipage}  
        \hfill
        \begin{minipage}[b]{0.119\linewidth}
            \centering
            \centerline{\includegraphics[width=\linewidth]{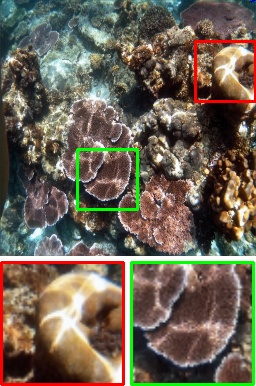}}
        \end{minipage}   
        \hfill       
        \begin{minipage}[b]{0.119\linewidth}
            \centering
            \centerline{\includegraphics[width=\linewidth]{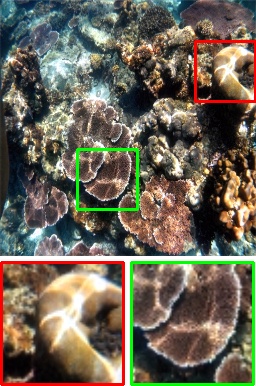}}
        \end{minipage}
        \hfill
        \begin{minipage}[b]{0.119\linewidth}
            \centering
            \centerline{\includegraphics[width=\linewidth]{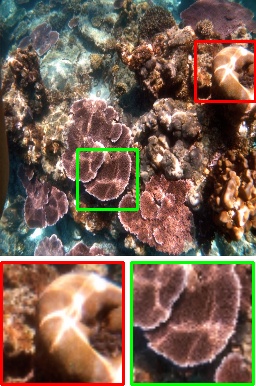}}
        \end{minipage}
        \hfill
        \begin{minipage}[b]{0.119\linewidth}
            \centering
            \centerline{\includegraphics[width=\linewidth]{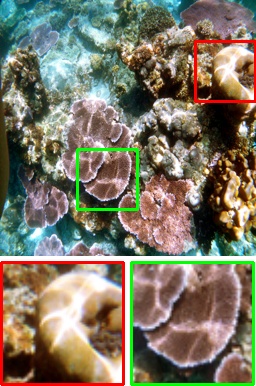}}
        \end{minipage}
    \end{minipage}
    
    \begin{minipage}[b]{1.0\linewidth}
        \begin{minipage}[b]{0.119\linewidth}
            \centering
            \centerline{\includegraphics[width=\linewidth]{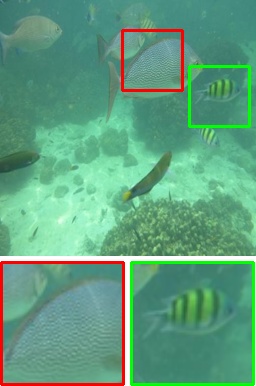}}
        \end{minipage}
         \hfill
        \begin{minipage}[b]{0.119\linewidth}
            \centering
            \centerline{\includegraphics[width=\linewidth]{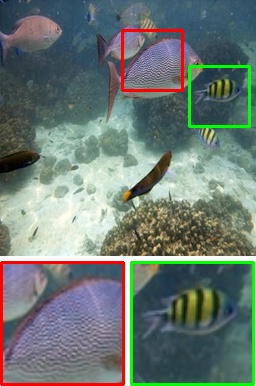}}
        \end{minipage}
        \hfill
        \begin{minipage}[b]{0.119\linewidth}
            \centering
            \centerline{\includegraphics[width=\linewidth]{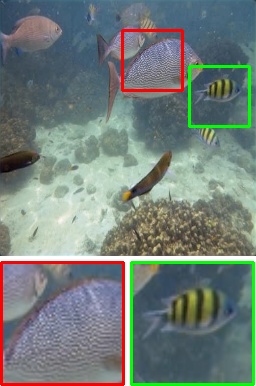}}
        \end{minipage}
        \hfill
        \begin{minipage}[b]{0.119\linewidth}
            \centering
            \centerline{\includegraphics[width=\linewidth]{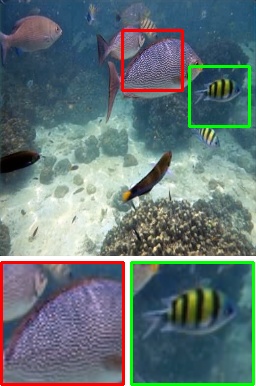}}
        \end{minipage}          
        \hfill
        \begin{minipage}[b]{0.119\linewidth}
            \centering
            \centerline{\includegraphics[width=\linewidth]{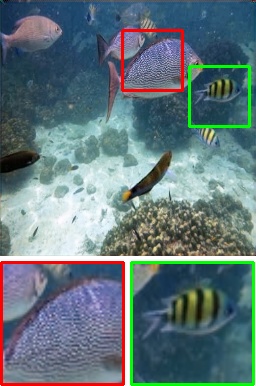}}
        \end{minipage}   
        \hfill
        \begin{minipage}[b]{0.119\linewidth}
            \centering
            \centerline{\includegraphics[width=\linewidth]{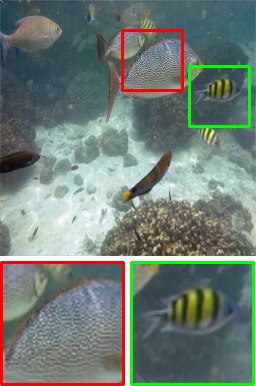}}
        \end{minipage}
        \hfill
        \begin{minipage}[b]{0.119\linewidth}
            \centering
            \centerline{\includegraphics[width=\linewidth]{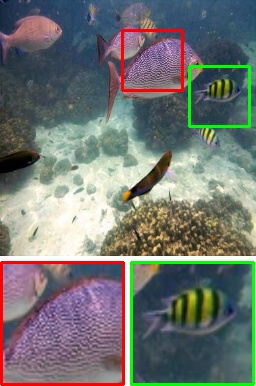}}
        \end{minipage}
        \hfill
        \begin{minipage}[b]{0.119\linewidth}
            \centering
            \centerline{\includegraphics[width=\linewidth]{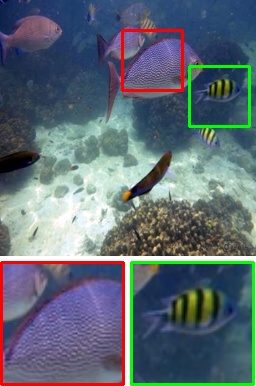}}
        \end{minipage}
    \end{minipage}
    
    \begin{minipage}[b]{1.0\linewidth}
        \begin{minipage}[b]{0.119\linewidth}
            \centering
            \centerline{\includegraphics[width=\linewidth]{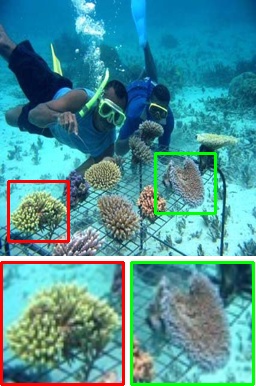}}
            \centerline{(a)Input}\medskip
        \end{minipage}   
        \hfill
        \begin{minipage}[b]{0.119\linewidth}
            \centering
            \centerline{\includegraphics[width=\linewidth]{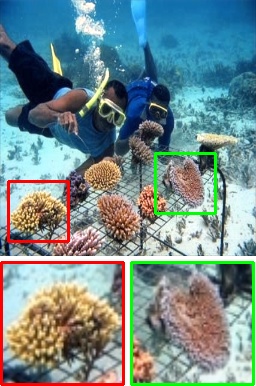}}
            \centerline{(b)URSCT}\medskip
        \end{minipage}
        \hfill
        \begin{minipage}[b]{0.119\linewidth}
            \centering
            \centerline{\includegraphics[width=\linewidth]{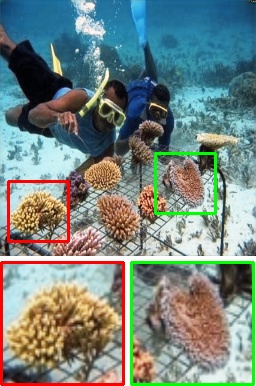}}
            \centerline{(c)TUDA}\medskip
        \end{minipage}   
        \hfill
        \begin{minipage}[b]{0.119\linewidth}
            \centering
            \centerline{\includegraphics[width=\linewidth]{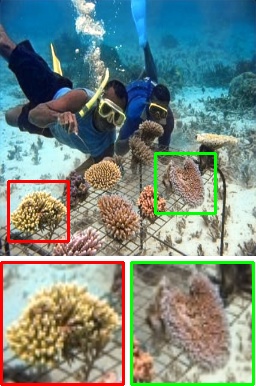}}
            \centerline{(d)WaterNet}\medskip
        \end{minipage}  
        \hfill
        \begin{minipage}[b]{0.119\linewidth}
            \centering
            \centerline{\includegraphics[width=\linewidth]{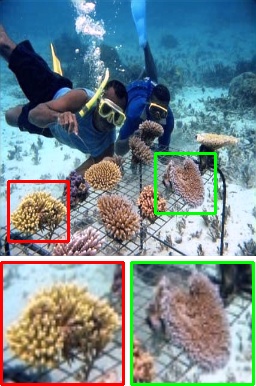}}
            \centerline{(e)GUPDM}\medskip
        \end{minipage}
        \hfill
        \begin{minipage}[b]{0.119\linewidth}
            \centering
            \centerline{\includegraphics[width=\linewidth]{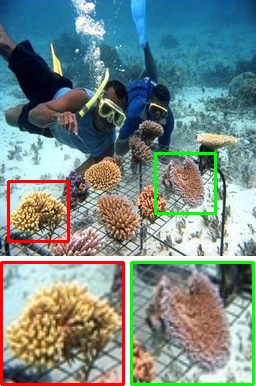}}
            \centerline{(f)Semi-UIR}\medskip
        \end{minipage}
        \hfill
        \begin{minipage}[b]{0.119\linewidth}
            \centering
            \centerline{\includegraphics[width=\linewidth]{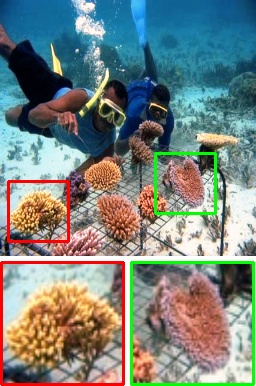}}
            \centerline{(g)Ours}\medskip
        \end{minipage}
        \hfill
        \begin{minipage}[b]{0.119\linewidth}
            \centering
            \centerline{\includegraphics[width=\linewidth]{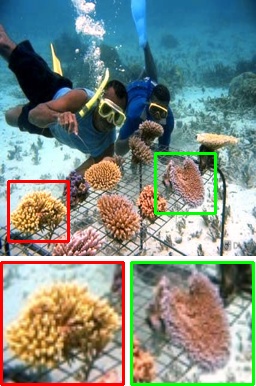}}
            \centerline{(h)Reference}\medskip
        \end{minipage}
    \end{minipage}
    \caption{
    Visual comparison with deep learning-based UIR methods. Upon inspecting the magnified sections, it is evident that our model markedly surpasses other methods in several aspects, including color saturation, contrast, and the richness and depth of colors. This superiority is observed in both well-lit and poorly-lit areas, with our model even exceeding the performance of the reference images presented in the last column. Additionally, our approach demonstrates superior performance in detail restoration, with notably clearer imagery in the red and green box area when compared to other methods.
    }
    \label{fig:dl-compare}
\end{figure*}

\subsection{Benchmark}
Existing UIR methods lack a standardized benchmark, hindering consistent comparisons due to disparities in training datasets, testing datasets, and evaluation metrics.

To address this issue and facilitate fair comparisons, we compiled a comprehensive benchmark using training data from three high-quality datasets known for their diversity and image quality: UIEB~\cite{li2019underwater}, EUVP~\cite{islam2020fast}, and LSUI~\cite{peng2023u}. The UIEB dataset offers higher-resolution images but includes only 890 paired images with references and 60 unpaired challenging images. Following the dataset's official guidelines, we randomly selected 800 paired images from UIEB for training. To enhance diversity while maintaining balanced sample sizes, we randomly selected 2,000 images each from the paired subsets of EUVP and LSUI. Although EUVP and LSUI contain over 11,000 and 4,000 paired images respectively, their original publications do not specify official train-test splits. We constructed our own balanced training set by selecting representative subsets, preventing any single dataset from dominating the training process and promoting a more generalized model. To further balance sample sizes, we duplicated the 800 UIEB training images, resulting in 1,600 samples from UIEB. The EUVP dataset, comprising three subsets labeled ``dark'', ``imagenet'', and ``scenes'', contributed 800, 700, and 500 images from each subset, respectively.

For evaluation, we assembled diverse paired and unpaired test data to assess performance and generalization. For the paired test sets with reference targets, we used the remaining 90 paired images from UIEB not included in the training set and selected 200 paired test samples each from EUVP and LSUI, ensuring no overlap with training data. Since official test sets for EUVP and LSUI are unspecified, we established standardized paired test sets by randomly selecting samples, ensuring consistent evaluation conditions and fair comparisons.

To assess generalization to untrained domains and challenging data, we included unpaired test sets without reference targets: 60 unpaired challenging images from UIEB and 200 unpaired test images each from EUVP and RUIE~\cite{liu2020real}. This diverse test data enables thorough evaluation across a wide range of underwater images with varying difficulty levels.

By designing our benchmark in this way, we aim to establish a standardized platform for evaluating UIR methods. Balancing training and testing data from multiple high-quality datasets ensures that models are trained and evaluated on diverse data without bias toward any single source, promoting fair assessments and facilitating meaningful comparisons.

\subsection{Evaluation Metrics}
\label{subsection:Metrics}
To evaluate underwater image restoration methods, we employ three commonly used reference-based metrics: Peak Signal-to-Noise Ratio (PSNR), Structural Similarity Index Measure (SSIM) \cite{wang2004image}, and Learned Perceptual Image Patch Similarity (LPIPS) \cite{zhang2018unreasonable}. Each metric assesses different aspects of image quality, providing a comprehensive evaluation of restoration performance. PSNR measures pixel-level accuracy, SSIM evaluates structural similarity, and LPIPS captures perceptual similarity. By employing these metrics alongside our benchmark, we ensure a thorough assessment of restored underwater images.

PSNR quantifies the ratio between the maximum possible power of a signal (an image) and the power of corrupting noise, with higher PSNR values indicating superior reconstruction quality. It is calculated as:
\begin{equation}
\text{PSNR} = 10 \cdot \log_{10} \left( \frac{MAX_I^2}{\text{MSE}} \right),
\label{eq:3}
\end{equation}
where $MAX_I$ is the maximum possible pixel value (255 for 8-bit images), and $\text{MSE}$ (Mean Squared Error) is calculated as:
\begin{equation}
\text{MSE} = \frac{1}{H \times W} \sum_{i=0}^{H-1} \sum_{j=0}^{W-1} [X(i, j) - Y(i, j)]^2,
\label{eq:4}
\end{equation}
with $H$ and $W$ representing the image dimensions, and $X(i,j)$ and $Y(i,j)$ being the pixel values of the original and restored images at position $(i,j)$, respectively.

SSIM assesses the structural similarity between the original and restored images based on luminance, contrast, and structural information, offering a more perceptually relevant evaluation. It is defined as:
\begin{equation}
\text{SSIM}(X, Y) = \frac{(2 \mu_X \mu_Y + C_1)(2 \sigma_{XY} + C_2)}{(\mu_X^2 + \mu_Y^2 + C_1)(\sigma_X^2 + \sigma_Y^2 + C_2)},
\label{eq:5}
\end{equation}
where $\mu_X$ and $\mu_Y$ are the mean pixel values of images $X$ and $Y$, $\sigma_X^2$ and $\sigma_Y^2$ are their variances, $\sigma_{XY}$ is the covariance, and $C_1$ and $C_2$ are constants to stabilize the division.

LPIPS is a perceptual metric that employs deep learning models to assess similarity by computing distances between feature representations at various network layers. A lower LPIPS score indicates greater perceptual similarity between the original and restored images.

Beyond these reference-based metrics, we incorporate two widely used non-reference metrics: Underwater Color Image Quality Evaluation (UCIQE) \cite{yang2015underwater} and Underwater Image Quality Measure (UIQM) \cite{panetta2015human}. These metrics evaluate underwater image quality without requiring a reference image. UCIQE considers attributes such as chroma, saturation, and contrast, while UIQM comprises measures of colorfulness, sharpness, and contrast as well.

However, despite their prevalence, UCIQE and UIQM have limitations in accurately reflecting the perceptual quality of enhanced underwater images, as noted in \cite{li2019underwater, guo2022underwater, liu2023uiqi} and corroborated by our own observations. As shown in Table \ref{tab:quan2}, many traditional methods achieve high UCIQE values, yet there is a significant discrepancy between these scores and the actual visual quality of the restored images, particularly for traditional methods' results. Similarly, methods with low full-reference metric results but high UIQM scores often produce images with poor visual quality. To illustrate these discrepancies, we have annotated the UCIQE and UIQM values beneath each result in Fig. \ref{fig:traditional-compare} and \ref{fig:nonref-compare}.

This discrepancy suggests that UCIQE and UIQM may not always accurately reflect the perceptual quality of restored underwater images, possibly due to their use of simplified human visual system (HVS) features combined in basic linear models, which neglect the non-linear nature of human perception of image quality \cite{guo2022underwater}. Consequently, methods adhering closely to certain graphical conventions may achieve high metric scores without yielding superior visual results.

Nevertheless, to assess generalization ability, we still employ UCIQE and UIQM as representative non-reference metrics, providing additional insights into the perceptual quality of restored images while acknowledging their limitations.

\subsection{Implementation Details}
The proposed model was developed utilizing the PyTorch framework and trained on an NVIDIA A100 GPU. The optimization algorithm employed was AdamW, with hyperparameters set to $\beta_1 = 0.9$ and $\beta_2 = 0.999$, and the training commenced with an initial learning rate of $2 \times 10^{-4}$. To modulate the learning rate effectively, a Cosine Annealing Learning Rate Scheduler (CosineAnnealingLR) was implemented. This scheduler modulates the learning rate in accordance with a cosine curve, progressively diminishing it from the initial value of $2 \times 10^{-4}$ to a nadir of $1 \times 10^{-6}$. 

The training process extended over 300 epochs with a batch size of 16. Input images were resized to a consistent resolution of $256 \times 256$ pixels. We also incorporated various data augmentation strategies, which included random cropping, flipping, rotation, transposition, and scaling of the images. 

\subsection{Comparisons with State-of-the-Art Methods}

To comprehensively evaluate our method against existing state-of-the-art underwater image restoration techniques, we compared our model with 37 representative methods, including 14 conventional and 23 deep learning based approaches. To ensure a fair and unbiased comparison, all deep learning methods were retrained on the same compiled benchmark dataset used in our study. The training settings, such as loss functions, number of iterations, and hyper-parameters, were kept consistent with those reported in their original papers.

\subsubsection{Quantitative Comparison}
To ensure a comprehensive evaluation, we assessed various methods on both paired and unpaired test sets. For the paired test sets—UIEB (90 images), EUVP (200 images), and LSUI (200 images)—we employed three full-reference metrics (PSNR, SSIM, LPIPS) and two non-reference metrics (UCIQE, UIQM).

Table \ref{tab:quan1} presents the results for the full-reference metrics. Our method consistently achieved the highest scores across most metrics and datasets. While ranking second to Semi-UIR in PSNR on the UIEB dataset by a marginal $0.04\%$, our method surpassed Semi-UIR significantly in SSIM ($3.9\%$ higher) and LPIPS ($16.7\%$ higher). This trend of outperforming Semi-UIR was consistent across all metrics on both the EUVP and LSUI datasets. Although URSCT scored slightly higher in SSIM on the EUVP and LSUI datasets ($0.93\%$ and $0.34\%$ respectively), our method excelled over URSCT in all other metrics across all three datasets. Notably, our method demonstrated a considerable advantage in LPIPS, which leverages deep learning models to better approximate human perception of image differences compared to traditional mathematical formulas. 

Table \ref{tab:quan2} presents the results for the non-reference metrics. While UCIQE and UIQM tends to assign high scores to the outputs from traditional methods, which often exhibit lower visual quality compared to deep learning approaches, we limit our non-reference comparison among the deep learning methods. While underwater-specific metrics like UCIQE and UIQM have limitations in reliably predicting the quality of enhanced underwater images, as noted in \cite{guo2022underwater} and through our own observations, they still offer some value as indicators of image attributes such as colorfulness, sharpness, and contrast. Our method achieved the highest UCIQE score on the UIEB test set and secured the second-highest scores on both the EUVP and LSUI datasets, indicating strong performance in chroma, contrast, and saturation. While our method did not achieve top rankings in UIQM, the results remained close to the average metric value, suggesting that our method meets the basic expectations for visual quality. 

To provide a holistic view of method efficiency, Table \ref{tab:quan1} also includes the computational cost in terms of MACs (Multiply-Accumulate Operations) and the number of parameters (Params) for each model. Although not the most efficient, our method maintains a lower computational cost than most other deep learning methods, demonstrating a favorable balance between performance and resource utilization.

\begin{figure*}[ht]
    \begin{minipage}[b]{1.0\linewidth}
        \begin{minipage}[b]{0.119\linewidth}
            \centering
            \centerline{\includegraphics[width=\linewidth]{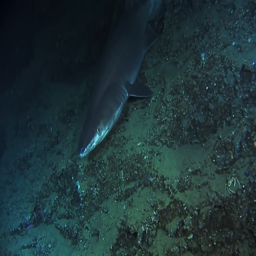}}
            \vspace{-2mm} \centerline{{\scriptsize UCIQE / UIQM}}
        \end{minipage}   
        \hfill
        \begin{minipage}[b]{0.119\linewidth}
            \centering
            \centerline{\includegraphics[width=\linewidth]{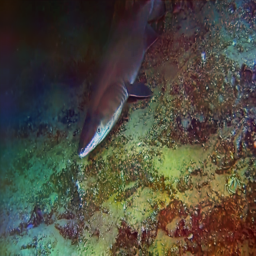}}
            \vspace{-2mm} \centerline{{\scriptsize 0.434 / \textbf{3.44}}}
        \end{minipage}
        \hfill
        \begin{minipage}[b]{0.119\linewidth}
            \centering
            \centerline{\includegraphics[width=\linewidth]{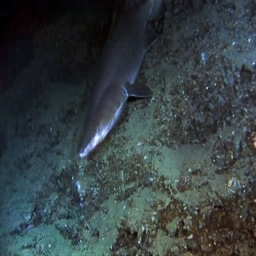}}
            \vspace{-2mm} \centerline{{\scriptsize 0.363 / 3.25}}
        \end{minipage}
        \hfill
        \begin{minipage}[b]{0.119\linewidth}
            \centering
            \centerline{\includegraphics[width=\linewidth]{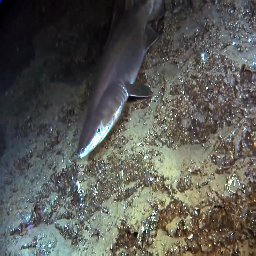}}
            \vspace{-2mm} \centerline{{\scriptsize 0.442 / 3.16}}
        \end{minipage}   
        \hfill
        \begin{minipage}[b]{0.119\linewidth}
            \centering
            \centerline{\includegraphics[width=\linewidth]{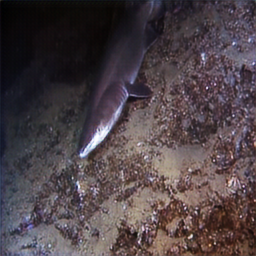}}
            \vspace{-2mm} \centerline{{\scriptsize \textbf{0.456} / 3.30}}
        \end{minipage}  
        \hfill
        \begin{minipage}[b]{0.119\linewidth}
            \centering
            \centerline{\includegraphics[width=\linewidth]{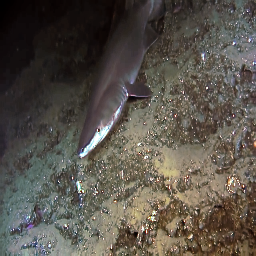}}
            \vspace{-2mm} \centerline{{\scriptsize 0.441 / 3.08}}
        \end{minipage}
        \hfill
        \begin{minipage}[b]{0.119\linewidth}
            \centering
            \centerline{\includegraphics[width=\linewidth]{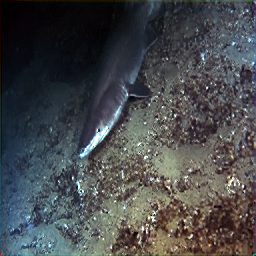}}
            \vspace{-2mm} \centerline{{\scriptsize 0.449 / 3.09}}
        \end{minipage}
        \hfill
        \begin{minipage}[b]{0.119\linewidth}
            \centering
            \centerline{\includegraphics[width=\linewidth]{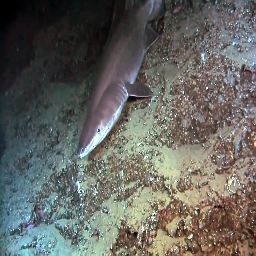}}
            \vspace{-2mm} \centerline{{\scriptsize 0.452 / 3.21}}
        \end{minipage}
    \end{minipage}

    \begin{minipage}[b]{1.0\linewidth}
        \begin{minipage}[b]{0.119\linewidth}
            \centering
            \centerline{\includegraphics[width=\linewidth]{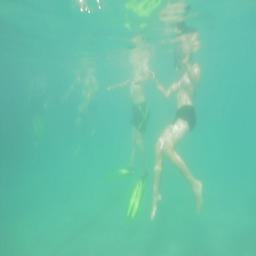}}
            \vspace{-2mm} \centerline{{\scriptsize UCIQE / UIQM}}
        \end{minipage}   
        \hfill
        \begin{minipage}[b]{0.119\linewidth}
            \centering
            \centerline{\includegraphics[width=\linewidth]{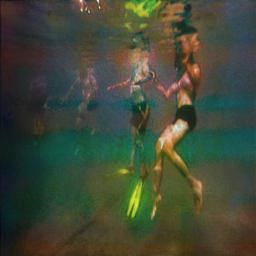}}
            \vspace{-2mm} \centerline{{\scriptsize 0.402 / \textbf{3.52}}}
        \end{minipage}
        \hfill
        \begin{minipage}[b]{0.119\linewidth}
            \centering
            \centerline{\includegraphics[width=\linewidth]{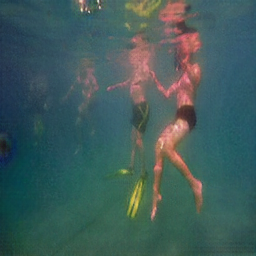}}
            \vspace{-2mm} \centerline{{\scriptsize 0.364 / 3.03}}
        \end{minipage}
        \hfill
        \begin{minipage}[b]{0.119\linewidth}
            \centering
            \centerline{\includegraphics[width=\linewidth]{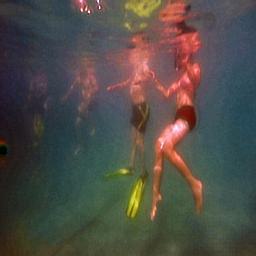}}
            \vspace{-2mm} \centerline{{\scriptsize \textbf{0.441} / 3.01}}
        \end{minipage}   
        \hfill
        \begin{minipage}[b]{0.119\linewidth}
            \centering
            \centerline{\includegraphics[width=\linewidth]{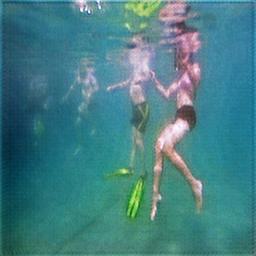}}
            \vspace{-2mm} \centerline{{\scriptsize 0.378 / 3.09}}
        \end{minipage}  
        \hfill
        \begin{minipage}[b]{0.119\linewidth}
            \centering
            \centerline{\includegraphics[width=\linewidth]{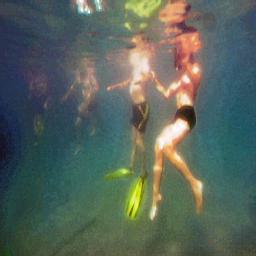}}
            \vspace{-2mm} \centerline{{\scriptsize 0.402 / 3.23}}
        \end{minipage}
        \hfill
        \begin{minipage}[b]{0.119\linewidth}
            \centering
            \centerline{\includegraphics[width=\linewidth]{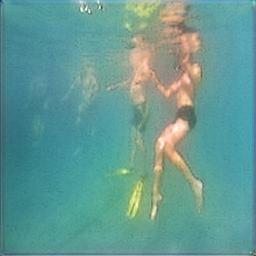}}
            \vspace{-2mm} \centerline{{\scriptsize 0.318 / 3.15}}
        \end{minipage}
        \hfill
        \begin{minipage}[b]{0.119\linewidth}
            \centering
            \centerline{\includegraphics[width=\linewidth]{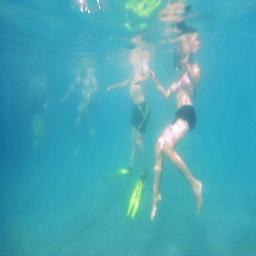}}
            \vspace{-2mm} \centerline{{\scriptsize 0.295 / 2.26}}
        \end{minipage}
    \end{minipage}

    \begin{minipage}[b]{1.0\linewidth}
        \begin{minipage}[b]{0.119\linewidth}
            \centering
            \centerline{\includegraphics[width=\linewidth]{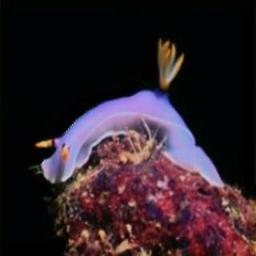}}
            \vspace{-2mm} \centerline{{\scriptsize UCIQE / UIQM}}
        \end{minipage}   
        \hfill
        \begin{minipage}[b]{0.119\linewidth}
            \centering
            \centerline{\includegraphics[width=\linewidth]{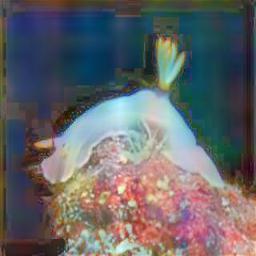}}
            \vspace{-2mm} \centerline{{\scriptsize 0.456 / \textbf{3.57}}}
        \end{minipage}
        \hfill
        \begin{minipage}[b]{0.119\linewidth}
            \centering
            \centerline{\includegraphics[width=\linewidth]{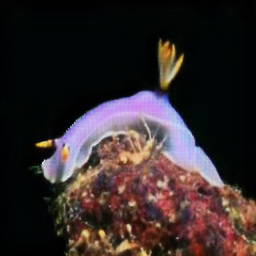}}
            \vspace{-2mm} \centerline{{\scriptsize 0.544 / 1.77}}
        \end{minipage}
        \hfill
        \begin{minipage}[b]{0.119\linewidth}
            \centering
            \centerline{\includegraphics[width=\linewidth]{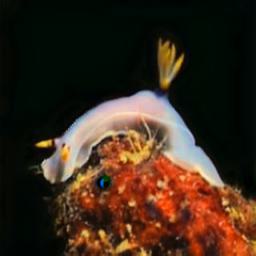}}
            \vspace{-2mm} \centerline{{\scriptsize 0.549 / 1.45}}
        \end{minipage}   
        \hfill
        \begin{minipage}[b]{0.119\linewidth}
            \centering
            \centerline{\includegraphics[width=\linewidth]{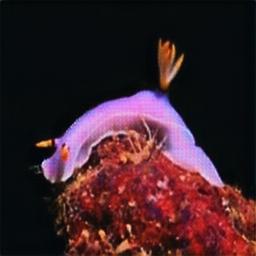}}
            \vspace{-2mm} \centerline{{\scriptsize 0.541 / 2.07}}
        \end{minipage}  
        \hfill
        \begin{minipage}[b]{0.119\linewidth}
            \centering
            \centerline{\includegraphics[width=\linewidth]{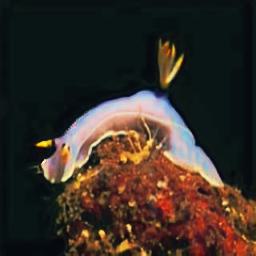}}
            \vspace{-2mm} \centerline{{\scriptsize 0.509 / 2.16}}
        \end{minipage}
        \hfill
        \begin{minipage}[b]{0.119\linewidth}
            \centering
            \centerline{\includegraphics[width=\linewidth]{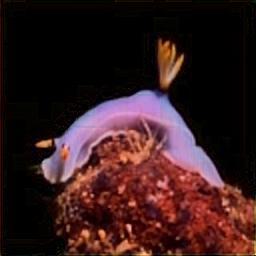}}
            \vspace{-2mm} \centerline{{\scriptsize \textbf{0.553} / 1.05}}
        \end{minipage}
        \hfill
        \begin{minipage}[b]{0.119\linewidth}
            \centering
            \centerline{\includegraphics[width=\linewidth]{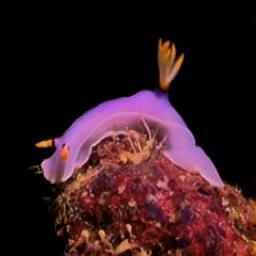}}
            \vspace{-2mm} \centerline{{\scriptsize 0.508 / 1.16}}
        \end{minipage}
    \end{minipage}

    \begin{minipage}[b]{1.0\linewidth}
        \begin{minipage}[b]{0.119\linewidth}
            \centering
            \centerline{\includegraphics[width=\linewidth]{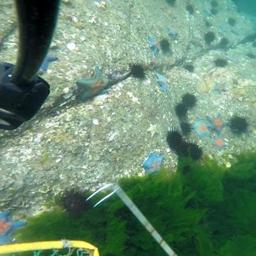}}
            \vspace{-2mm} \centerline{{\scriptsize UCIQE / UIQM}}
            \centerline{{\footnotesize (a)Input}}\medskip
        \end{minipage}   
        \hfill
        \begin{minipage}[b]{0.119\linewidth}
            \centering
            \centerline{\includegraphics[width=\linewidth]{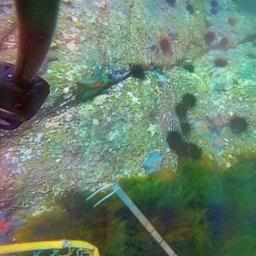}}
            \vspace{-2mm} \centerline{{\scriptsize 0.376 / \textbf{3.75}}}
            \centerline{{\footnotesize (b)CCMSRNet}}\medskip
        \end{minipage}
        \hfill
        \begin{minipage}[b]{0.119\linewidth}
            \centering
            \centerline{\includegraphics[width=\linewidth]{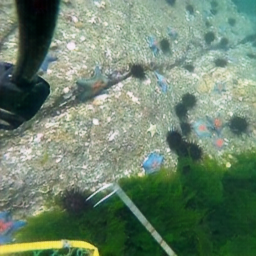}}
            \vspace{-2mm} \centerline{{\scriptsize 0.381 / 3.50}}
            \centerline{{\footnotesize (c)TACL}}\medskip
        \end{minipage}
        \hfill
        \begin{minipage}[b]{0.119\linewidth}
            \centering
            \centerline{\includegraphics[width=\linewidth]{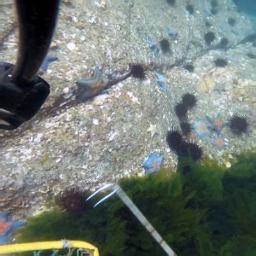}}
            \vspace{-2mm} \centerline{{\scriptsize 0.412 / 3.47}}
            \centerline{{\footnotesize (d)SyreaNet}}\medskip
        \end{minipage}   
        \hfill
        \begin{minipage}[b]{0.119\linewidth}
            \centering
            \centerline{\includegraphics[width=\linewidth]{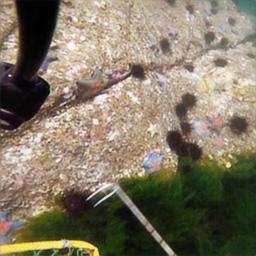}}
            \vspace{-2mm} \centerline{{\scriptsize 0.441 / 3.28}}
            \centerline{{\footnotesize (e)UGAN}}\medskip
        \end{minipage}  
        \hfill
        \begin{minipage}[b]{0.119\linewidth}
            \centering
            \centerline{\includegraphics[width=\linewidth]{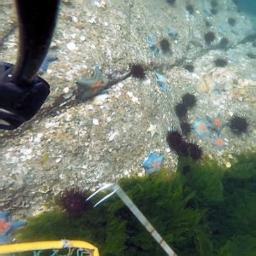}}
            \vspace{-2mm} \centerline{{\scriptsize 0.410 / 3.50}}
            \centerline{{\footnotesize (f)USUIR}}\medskip
        \end{minipage}
        \hfill
        \begin{minipage}[b]{0.119\linewidth}
            \centering
            \centerline{\includegraphics[width=\linewidth]{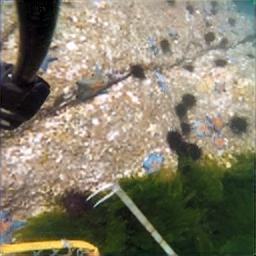}}
            \vspace{-2mm} \centerline{{\scriptsize 0.432 / 3.40}}
            \centerline{{\footnotesize (g)U-Transformer}}\medskip
        \end{minipage}
        \hfill
        \begin{minipage}[b]{0.119\linewidth}
            \centering
            \centerline{\includegraphics[width=\linewidth]{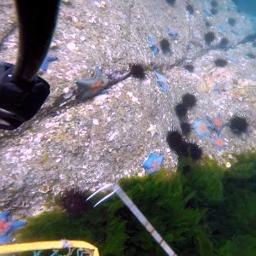}}
            \vspace{-2mm} \centerline{{\scriptsize \textbf{0.474} / 3.28}}
            \centerline{{\footnotesize (h)Ours}}\medskip
        \end{minipage}
    \end{minipage}

    \caption{Visual comparison of our method with the top 6 deep learning methods based on average UCIQE scores on three unpaired test sets: UIEB, EUVP, and RUIE. Despite our method ranking third in terms of average UCIQE score among the deep learning approaches, the visual quality of the images restored by our method is superior, showcasing its effectiveness in enhancing underwater images across diverse datasets and challenging scenarios. }
    
    \label{fig:nonref-compare}
\end{figure*}
\begin{table}[]
\centering
\caption{Quantitative comparison of deep learning methods on the unpaired test sets UIEB, EUVP, and RUIE without reference targets using the \textbf{UCIQE} metric.}
\adjustbox{width=\columnwidth}{%
\begin{tabular}{lrrrlrrr}
\toprule
\textbf{Method}        & \textbf{UIEB} & \textbf{EUVP} & \textbf{RUIE} & \textbf{Method}          & \textbf{UIEB}  & \textbf{EUVP}  & \textbf{RUIE}  \\
\midrule
\textbf{UGAN}          & 0.427         & 0.425         & 0.407         & \textbf{CCMSRNet}        & 0.414          & \underline{0.435}    & \textbf{0.436} \\
\textbf{WaterNet}      & 0.426         & 0.419         & 0.393         & \textbf{DeepWaveNet}     & 0.413          & 0.395          & 0.365          \\
\textbf{UWCNN}         & 0.354         & 0.319         & 0.255         & \textbf{GUPDM}           & 0.412          & 0.382          & 0.380          \\
\textbf{Shallow-UWNet} & 0.337         & 0.316         & 0.240         & \textbf{MBANet}          & 0.409          & 0.400          & 0.380          \\
\textbf{Ucolor}        & 0.352         & 0.322         & 0.256         & \textbf{$ \mathbf{NU^2Net}$} & 0.383          & 0.390          & 0.355          \\
\textbf{CLUIE-Net}     & 0.394         & 0.349         & 0.331         & \textbf{PUGAN}           & 0.414          & 0.412          & 0.369          \\
\textbf{PUIE-Net}      & 0.354         & 0.349         & 0.336         & \textbf{Semi-UIR}        & 0.417          & 0.400          & 0.385          \\
\textbf{STSC}          & 0.393         & 0.385         & 0.378         & \textbf{SyreaNet}        & \textbf{0.433} & \textbf{0.436} & 0.391          \\
\textbf{TACL}          & 0.421         & 0.432         & 0.419         & \textbf{TUDA}            & 0.422          & 0.405          & 0.392          \\
\textbf{UIE-WD}        & 0.403         & 0.380         & 0.343         & \textbf{U-Transformer}   & 0.424          & 0.406          & 0.409          \\
\textbf{URSCT}         & 0.414         & 0.398         & 0.384         & \textbf{SFGNet}          & 0.412          & 0.388          & 0.390          \\
\textbf{USUIR}         & \underline{0.431}   & 0.422         & 0.390         & \textbf{Ours}            & \underline{0.431}    & 0.407          & \underline{0.426}   \\
\bottomrule
\end{tabular}
}
\label{tab:uciqe_nonref}
\end{table}
For the unpaired evaluation, we employed the non-reference metric UCIQE to assess performance on the UIEB (60 images), EUVP (200 images), and RUIE (200 images) datasets. Given the previously discussed discrepancies between high UCIQE scores and actual visual quality in traditional methods, our analysis focused solely on deep learning approaches. Table \ref{tab:uciqe_nonref} presents the UCIQE results for these deep learning methods. Although our method did not attain the highest overall score, its consistently high ranking across the datasets underscores its robust performance and capacity for effective generalization across diverse domains.

All these quantitative results clearly highlights the outstanding effectiveness of our GuidedHybSensUIR.

\subsubsection{Qualitative Comparison}
The visual comparisons between our method and SOTA methods, which include the top 5 traditional and top 5 deep learning methods based on average metric data in Table \ref{tab:quan1}, are illustrated in Fig. \ref{fig:traditional-compare} and Fig. \ref{fig:dl-compare}. For a better view, please zoom in.
The observations from Fig. \ref{fig:traditional-compare} indicate that traditional methods often falter in complex ocean environments. They frequently fail to eliminate the haze resulting from underwater light attenuation and turbidity, or they might introduce new color distortions. Moreover, issues of over- or under-exposure are commonplace, leading to colors that appear unrealistic. In contrast, our method consistently excels in correcting distorted colors and restoring blurred details over large distances.

The overall performance of deep learning-based UIR methods, particularly those with high quantitative metrics, significantly outperforms traditional approaches. Their qualitative superiority is showcased in Fig. \ref{fig:dl-compare}. To better highlight our method's superiority, we magnified local areas that exhibit richer color differences and textual details. It is evident that our method achieves improved color saturation and contrast, resulting in more vibrant overall colors. Additionally, the details within the red box and green box illustrate our method's capability to restore details more effectively. This restoration is evident not only in the detailed texture but also in the fine-grained color. Overall, our method outperforms state-of-the-art methods in various aspects, including both highlight and dark areas, and in terms of both color and texture details.

Furthermore, a qualitative comparison with deep learning methods that achieved high UCIQE scores, the top 6 average UCIQE scores in Table \ref{tab:uciqe_nonref}, on three unpaired test sets (UIEB, EUVP, and RUIE) is shown in Fig. \ref{fig:nonref-compare}. This comparison demonstrates that our method's actual generalization performance is superior compared to any of the other methods. Our approach showcases robustness and adaptability in handling underwater images from various sources and with different characteristics, underlining its potential for real-world applications.

\subsection{Ablation Studies}
\begin{table*}[ht]
\centering
\caption{Ablation study results averaged across the UIEB, EUVP, and LSUI test sets. All models were trained on the benchmark dataset presented in this paper, maintaining consistency with the full GuidedHybSensUIR model and other compared methods.}
\adjustbox{width=0.8\textwidth}{%
\begin{tabular}{ccccccccccrrrr}
\toprule
\multirow{2}{*}{Backbone} & \multicolumn{2}{c}{Prior} & \multicolumn{3}{c}{FC} & \multicolumn{2}{c}{DR} & \multirow{2}{*}{SH} & \multirow{2}{*}{Skip Connection} & \multirow{2}{*}{PSNR$\uparrow$} & \multirow{2}{*}{SSIM$\uparrow$} & \multirow{2}{*}{LPIPS$\downarrow$} & \multirow{2}{*}{UCIQE$\uparrow$} \\
\cmidrule(lr){2-3} \cmidrule(lr){4-6} \cmidrule(lr){7-8}
                          & guide FC    & Skip Connection   & SAT    & ACT   & KFT   & RCB       & NAFB       &                     &                            &                        &                        &                         &                         \\
\midrule
\Checkmark                         & \tiny{\XSolidBold}           & \tiny{\XSolidBold}           & \tiny{\XSolidBold}      & \tiny{\XSolidBold}     & \tiny{\XSolidBold}     & \tiny{\XSolidBold}         & \tiny{\XSolidBold}          & \tiny{\XSolidBold}                   & \tiny{\XSolidBold}                          & 22.91                  & 0.850                  & 0.150                   & 0.403                   \\
\Checkmark                         & \tiny{\XSolidBold}           & \tiny{\XSolidBold}           & \Checkmark      & \tiny{\XSolidBold}     & \tiny{\XSolidBold}     & \tiny{\XSolidBold}         & \tiny{\XSolidBold}          & \tiny{\XSolidBold}                   & \tiny{\XSolidBold}                          & 23.57                  & 0.853                  & 0.139                   & 0.403                   \\
\Checkmark                         & \Checkmark           & \tiny{\XSolidBold}           & \Checkmark      & \Checkmark     & \tiny{\XSolidBold}     & \tiny{\XSolidBold}         & \tiny{\XSolidBold}          & \tiny{\XSolidBold}                   & \tiny{\XSolidBold}                          & 23.86                  & 0.856                  & 0.137                   & 0.421                   \\
\Checkmark                         & \Checkmark           & \tiny{\XSolidBold}           & \Checkmark      & \Checkmark     & \Checkmark     & \tiny{\XSolidBold}         & \tiny{\XSolidBold}          & \tiny{\XSolidBold}                   & \tiny{\XSolidBold}                          & 24.22                  & 0.861                  & 0.124                   & 0.412                   \\
\Checkmark                         & \tiny{\XSolidBold}           & \tiny{\XSolidBold}           & \tiny{\XSolidBold}      & \tiny{\XSolidBold}     & \tiny{\XSolidBold}     & \Checkmark         & \tiny{\XSolidBold}          & \tiny{\XSolidBold}                   & \tiny{\XSolidBold}                          & 24.05                  & 0.856                  & 0.122                   & 0.408                   \\
\Checkmark                         & \tiny{\XSolidBold}           & \tiny{\XSolidBold}           & \tiny{\XSolidBold}      & \tiny{\XSolidBold}     & \tiny{\XSolidBold}     & \Checkmark         & \Checkmark          & \tiny{\XSolidBold}                   & \tiny{\XSolidBold}                          & 24.15                  & 0.861                  & 0.116                   & 0.420                   \\
\Checkmark                         & \Checkmark           & \tiny{\XSolidBold}           & \Checkmark      & \Checkmark     & \Checkmark     & \Checkmark         & \Checkmark          & \tiny{\XSolidBold}                   & \tiny{\XSolidBold}                          & 24.72                  & 0.874                  & 0.110                   & 0.430                   \\
\Checkmark                         & \tiny{\XSolidBold}           & \tiny{\XSolidBold}           & \tiny{\XSolidBold}      & \tiny{\XSolidBold}     & \tiny{\XSolidBold}     & \tiny{\XSolidBold}         & \tiny{\XSolidBold}          & \Checkmark                   & \tiny{\XSolidBold}                          & 23.83                  & 0.869                  & 0.142                   & 0.410                   \\
\Checkmark                         & \Checkmark           & \tiny{\XSolidBold}           & \Checkmark      & \Checkmark     & \Checkmark     & \Checkmark         & \Checkmark          & \Checkmark                   & \tiny{\XSolidBold}                          & 24.99                  & 0.880                  & 0.109                   & 0.431                   \\
\Checkmark                         & \Checkmark           & \Checkmark           & \Checkmark      & \Checkmark     & \Checkmark     & \Checkmark         & \Checkmark          & \Checkmark                   & \tiny{\XSolidBold}                          & 25.08                  & 0.882                  & 0.103                   & 0.425                   \\
\Checkmark                         & \Checkmark           & \Checkmark           & \Checkmark      & \Checkmark     & \Checkmark     & \Checkmark         & \Checkmark          & \Checkmark                   & \Checkmark                          & 25.25                  & 0.883                  & 0.103                   & 0.435              \\    
\bottomrule
\end{tabular}
}
\label{tab:ablation}
\end{table*}
To validate the contribution of each component within our proposed GuidedHybSensUIR architecture, we conducted a comprehensive ablation study. Table \ref{tab:ablation} presents the results averaged across the UIEB, EUVP, and LSUI test sets. All models were trained on our benchmark dataset to ensure consistency. Starting with a backbone UNet having the same downsampling depth, architecture, and number of modules as our full model — but with a higher parameter count (1.469 million compared to our model's 1.145 million) — we incrementally replaced the original UNet modules with our proposed modules at corresponding positions. This approach ensures that performance gains are not due to increased parameter counts.

Our model comprises Detail Restorers (DR) in the encoder, a Feature Contextualizer (FC) at the bottleneck, and Scale Harmonizers (SH) in the decoder. The Color Balance Prior is embedded within the FC and skip-connected to the decoder output. The DR module consists of Residual Context Block (RCB) and Nonlinear Activation-Free Block (NAFB), while the FC includes Adjust Color Transformer (ACT), Keep Feature Transformer (KFT), and Self-Attention Transformer (SAT). We conducted ten ablation experiments to assess the individual and combined effectiveness of these components.

We first validated individual components within the Feature Contextualizer (rows 2–4 in Table \ref{tab:ablation}), gradually adding them to the backbone UNet. The SAT showed the most substantial impact in capturing self-attention. Since the ACT and KFT focus on cross-attention between image features and prior features, we integrated the corresponding prior branch when evaluating them. Both ACT and KFT brought significant improvements, highlighting the importance of modeling long-range dependencies and global color relationships.

Both the RCB and NAFB within the Detail Restorer contributed positively to performance. Integrating both DR and FC with the backbone (row 7) yielded significant improvements across all metrics compared to using FC alone (row 4), with substantial gains in PSNR and SSIM, emphasizing the DR module's role in recovering fine-grained details. Including Scale Harmonizer (SH) modules further improved results, underscoring their effectiveness in integrating multi-scale features.

Incorporating the color balance prior with its associated skip connection (``skip conn'' under ``Prior'') significantly improved performance over the baseline, demonstrating its effectiveness in guiding accurate color restoration. Adding a skip connection (``skip conn'') between the input image and final output consistently improved results, showcasing the benefits of preserving information and facilitating multi-scale feature fusion.

The best overall performance was achieved when all components were enabled, highlighting the synergistic effect of our proposed GuidedHybSensUIR architecture in combining global and local enhancements, skip connections, and a color balance prior to achieve state-of-the-art underwater image restoration.

\section{Conclusion}
In this paper, we introduce the Hybrid Sense Underwater Image Restoration method, innovatively guided by a Color Balance Prior. This method employs a Detail Restorer, adept at restoring fine image details across various scales using CNN, renowned for its local feature extraction capabilities. For broader scale details, we leverage Transformer, which uniquely applies inter-channel attention to capture long-range relationships, attending both cross-attention and self-attention, thereby enhancing color performance. A specially designed Scale Harmonizer is implemented to effectively merge features from different scales. Crucially, our Color Balance Prior guides the model towards a more stable and effective trajectory, aiming for the global optimizer. Additionally, we compiled a new benchmark and conducted extensive comparative experiments on it, laying a solid foundation for future advancements in this field. Our comprehensive quantitative and qualitative analyses demonstrate that our method surpasses state-of-the-art techniques in underwater image restoration. 


\bibliographystyle{IEEEtran}
\bibliography{IEEEabrv,ref}

\begin{thebibliography}{10}
\providecommand{\url}[1]{#1}
\csname url@samestyle\endcsname
\providecommand{\newblock}{\relax}
\providecommand{\bibinfo}[2]{#2}
\providecommand{\BIBentrySTDinterwordspacing}{\spaceskip=0pt\relax}
\providecommand{\BIBentryALTinterwordstretchfactor}{4}
\providecommand{\BIBentryALTinterwordspacing}{\spaceskip=\fontdimen2\font plus
\BIBentryALTinterwordstretchfactor\fontdimen3\font minus \fontdimen4\font\relax}
\providecommand{\BIBforeignlanguage}[2]{{%
\expandafter\ifx\csname l@#1\endcsname\relax
\typeout{** WARNING: IEEEtran.bst: No hyphenation pattern has been}%
\typeout{** loaded for the language `#1'. Using the pattern for}%
\typeout{** the default language instead.}%
\else
\language=\csname l@#1\endcsname
\fi
#2}}
\providecommand{\BIBdecl}{\relax}
\BIBdecl

\bibitem{li2021marine}
L.~Li, B.~Dong, E.~Rigall, T.~Zhou, J.~Dong, and G.~Chen, ``Marine animal segmentation,'' \emph{IEEE Transactions on Circuits and Systems for Video Technology}, vol.~32, no.~4, pp. 2303--2314, 2021.

\bibitem{zhao2018weakly}
F.~Zhao, J.~Li, J.~Zhao, and J.~Feng, ``Weakly supervised phrase localization with multi-scale anchored transformer network,'' in \emph{Proceedings of the IEEE Conference on Computer Vision and Pattern Recognition}, 2018, pp. 5696--5705.

\bibitem{zhao2019look}
J.~Zhao, Y.~Cheng, Y.~Cheng, Y.~Yang, F.~Zhao, J.~Li, H.~Liu, S.~Yan, and J.~Feng, ``Look across elapse: Disentangled representation learning and photorealistic cross-age face synthesis for age-invariant face recognition,'' in \emph{Proceedings of the AAAI conference on artificial intelligence}, vol.~33, no.~01, 2019, pp. 9251--9258.

\bibitem{tu2021joint}
X.~Tu, J.~Zhao, Q.~Liu, W.~Ai, G.~Guo, Z.~Li, W.~Liu, and J.~Feng, ``Joint face image restoration and frontalization for recognition,'' \emph{IEEE Transactions on circuits and systems for video technology}, vol.~32, no.~3, pp. 1285--1298, 2021.

\bibitem{liu2022rank}
J.~Liu, R.~W. Liu, J.~Sun, and T.~Zeng, ``Rank-one prior: Real-time scene recovery,'' \emph{IEEE Transactions on Pattern Analysis and Machine Intelligence}, 2022.

\bibitem{zhang2023underwater}
W.~Zhang, L.~Zhou, P.~Zhuang, G.~Li, X.~Pan, W.~Zhao, and C.~Li, ``Underwater image enhancement via weighted wavelet visual perception fusion,'' \emph{IEEE Transactions on Circuits and Systems for Video Technology}, 2023.

\bibitem{wang2023domain}
Z.~Wang, L.~Shen, M.~Xu, M.~Yu, K.~Wang, and Y.~Lin, ``Domain adaptation for underwater image enhancement,'' \emph{IEEE Transactions on Image Processing}, vol.~32, pp. 1442--1457, 2023.

\bibitem{wen2023syreanet}
J.~Wen, J.~Cui, Z.~Zhao, R.~Yan, Z.~Gao, L.~Dou, and B.~M. Chen, ``Syreanet: A physically guided underwater image enhancement framework integrating synthetic and real images,'' in \emph{IEEE International Conference on Robotics and Automation}, 2023.

\bibitem{huang2023contrastive}
S.~Huang, K.~Wang, H.~Liu, J.~Chen, and Y.~Li, ``Contrastive semi-supervised learning for underwater image restoration via reliable bank,'' in \emph{Computer Vision and Pattern Recognition}, 2023, pp. 18\,145--18\,155.

\bibitem{xue2023investigating}
X.~Xue, Z.~Li, L.~Ma, Q.~Jia, R.~Liu, and X.~Fan, ``Investigating intrinsic degradation factors by multi-branch aggregation for real-world underwater image enhancement,'' \emph{Pattern Recognition}, vol. 133, p. 109041, 2023.

\bibitem{peng2023u}
L.~Peng, C.~Zhu, and L.~Bian, ``U-shape transformer for underwater image enhancement,'' \emph{IEEE Transactions on Image Processing}, 2023.

\bibitem{dosovitskiy2020image}
A.~Dosovitskiy, L.~Beyer, A.~Kolesnikov, D.~Weissenborn, X.~Zhai, T.~Unterthiner, M.~Dehghani, M.~Minderer, G.~Heigold, S.~Gelly, J.~Uszkoreit, and N.~Houlsby, ``An image is worth 16x16 words: Transformers for image recognition at scale,'' \emph{International Conference on Learning Representations}, 2021.

\bibitem{buchsbaum1980spatial}
G.~Buchsbaum, ``A spatial processor model for object colour perception,'' \emph{Journal of the Franklin institute}, vol. 310, no.~1, pp. 1--26, 1980.

\bibitem{mu2023generalized}
P.~Mu, H.~Xu, Z.~Liu, Z.~Wang, S.~Chan, and C.~Bai, ``A generalized physical-knowledge-guided dynamic model for underwater image enhancement,'' in \emph{ACM International Conference on Multimedia}, 2023, pp. 7111--7120.

\bibitem{li2021underwater}
C.~Li, S.~Anwar, J.~Hou, R.~Cong, C.~Guo, and W.~Ren, ``Underwater image enhancement via medium transmission-guided multi-color space embedding,'' \emph{IEEE Transactions on Image Processing}, vol.~30, pp. 4985--5000, 2021.

\bibitem{li2019underwater}
C.~Li, C.~Guo, W.~Ren, R.~Cong, J.~Hou, S.~Kwong, and D.~Tao, ``An underwater image enhancement benchmark dataset and beyond,'' \emph{IEEE Transactions on Image Processing}, vol.~29, pp. 4376--4389, 2019.

\bibitem{islam2020fast}
M.~J. Islam, Y.~Xia, and J.~Sattar, ``Fast underwater image enhancement for improved visual perception,'' \emph{IEEE Robotics and Automation Letters}, vol.~5, no.~2, pp. 3227--3234, 2020.

\bibitem{liu2020real}
R.~Liu, X.~Fan, M.~Zhu, M.~Hou, and Z.~Luo, ``Real-world underwater enhancement: Challenges, benchmarks, and solutions under natural light,'' \emph{IEEE Transactions on Circuits and Systems for Video Technology}, vol.~30, no.~12, pp. 4861--4875, 2020.

\bibitem{he2010single}
K.~He, J.~Sun, and X.~Tang, ``Single image haze removal using dark channel prior,'' \emph{IEEE Transactions on Pattern Analysis and Machine Intelligence}, vol.~33, no.~12, pp. 2341--2353, 2010.

\bibitem{drews2013transmission}
P.~Drews, E.~Nascimento, F.~Moraes, S.~Botelho, and M.~Campos, ``Transmission estimation in underwater single images,'' in \emph{International Conference on Computer Vision Workshop}, 2013, pp. 825--830.

\bibitem{li2016single}
C.~Li, J.~Quo, Y.~Pang, S.~Chen, and J.~Wang, ``Single underwater image restoration by blue-green channels dehazing and red channel correction,'' in \emph{IEEE International Conference on Acoustics, Speech, and Signal Processing}, 2016, pp. 1731--1735.

\bibitem{peng2018generalization}
Y.-T. Peng, K.~Cao, and P.~C. Cosman, ``Generalization of the dark channel prior for single image restoration,'' \emph{IEEE Transactions on Image Processing}, vol.~27, no.~6, pp. 2856--2868, 2018.

\bibitem{xie2021variational}
J.~Xie, G.~Hou, G.~Wang, and Z.~Pan, ``A variational framework for underwater image dehazing and deblurring,'' \emph{IEEE Transactions on Circuits and Systems for Video Technology}, vol.~32, no.~6, pp. 3514--3526, 2021.

\bibitem{zhou2021underwater}
J.~Zhou, Z.~Liu, W.~Zhang, D.~Zhang, and W.~Zhang, ``Underwater image restoration based on secondary guided transmission map,'' \emph{Multimedia Tools and Applications}, vol.~80, pp. 7771--7788, 2021.

\bibitem{chiang2011underwater}
J.~Y. Chiang and Y.-C. Chen, ``Underwater image enhancement by wavelength compensation and dehazing,'' \emph{IEEE Transactions on Image Processing}, vol.~21, no.~4, pp. 1756--1769, 2011.

\bibitem{akkaynak2019sea}
D.~Akkaynak and T.~Treibitz, ``Sea-thru: A method for removing water from underwater images,'' in \emph{Computer Vision and Pattern Recognition}, 2019, pp. 1682--1691.

\bibitem{zhou2023underwater}
J.~Zhou, Q.~Liu, Q.~Jiang, W.~Ren, K.-M. Lam, and W.~Zhang, ``Underwater camera: Improving visual perception via adaptive dark pixel prior and color correction,'' \emph{International Journal of Computer Vision}, pp. 1--19, 2023.

\bibitem{song2020enhancement}
W.~Song, Y.~Wang, D.~Huang, A.~Liotta, and C.~Perra, ``Enhancement of underwater images with statistical model of background light and optimization of transmission map,'' \emph{IEEE Transactions on Broadcasting}, vol.~66, no.~1, pp. 153--169, 2020.

\bibitem{li2016underwater}
C.-Y. Li, J.-C. Guo, R.-M. Cong, Y.-W. Pang, and B.~Wang, ``Underwater image enhancement by dehazing with minimum information loss and histogram distribution prior,'' \emph{IEEE Transactions on Image Processing}, vol.~25, no.~12, pp. 5664--5677, 2016.

\bibitem{zhang2022underwater}
W.~Zhang, P.~Zhuang, H.-H. Sun, G.~Li, S.~Kwong, and C.~Li, ``Underwater image enhancement via minimal color loss and locally adaptive contrast enhancement,'' \emph{IEEE Transactions on Image Processing}, vol.~31, pp. 3997--4010, 2022.

\bibitem{peng2017underwater}
Y.-T. Peng and P.~C. Cosman, ``Underwater image restoration based on image blurriness and light absorption,'' \emph{IEEE Transactions on Image Processing}, vol.~26, no.~4, pp. 1579--1594, 2017.

\bibitem{berman2017diving}
D.~Berman, T.~Treibitz, and S.~Avidan, ``Diving into haze-lines: Color restoration of underwater images,'' in \emph{British Machine Vision Conference}, vol.~1, no.~2, 2017, p.~2.

\bibitem{berman2020underwater}
D.~Berman, D.~Levy, S.~Avidan, and T.~Treibitz, ``Underwater single image color restoration using haze-lines and a new quantitative dataset,'' \emph{IEEE Transactions on Pattern Analysis and Machine Intelligence}, vol.~43, no.~8, pp. 2822--2837, 2020.

\bibitem{hou2023non}
G.~Hou, N.~Li, P.~Zhuang, K.~Li, H.~Sun, and C.~Li, ``Non-uniform illumination underwater image restoration via illumination channel sparsity prior,'' \emph{IEEE Transactions on Circuits and Systems for Video Technology}, 2023.

\bibitem{ancuti2012enhancing}
C.~Ancuti, C.~O. Ancuti, T.~Haber, and P.~Bekaert, ``Enhancing underwater images and videos by fusion,'' in \emph{Computer Vision and Pattern Recognition}, 2012, pp. 81--88.

\bibitem{jaffe1990computer}
J.~S. Jaffe, ``Computer modeling and the design of optimal underwater imaging systems,'' \emph{IEEE Journal of Oceanic Engineering}, vol.~15, no.~2, pp. 101--111, 1990.

\bibitem{hitam2013mixture}
M.~S. Hitam, E.~A. Awalludin, W.~N. J. H.~W. Yussof, and Z.~Bachok, ``Mixture contrast limited adaptive histogram equalization for underwater image enhancement,'' in \emph{ICCAT}, 2013, pp. 1--5.

\bibitem{iqbal2010enhancing}
K.~Iqbal, M.~Odetayo, A.~James, R.~A. Salam, and A.~Z.~H. Talib, ``Enhancing the low quality images using unsupervised colour correction method,'' in \emph{IEEE SMC}, 2010, pp. 1703--1709.

\bibitem{zhang2017underwater}
S.~Zhang, T.~Wang, J.~Dong, and H.~Yu, ``Underwater image enhancement via extended multi-scale retinex,'' \emph{Neurocomputing}, vol. 245, pp. 1--9, 2017.

\bibitem{zhuang2021bayesian}
P.~Zhuang, C.~Li, and J.~Wu, ``Bayesian retinex underwater image enhancement,'' \emph{Engineering Applications of Artificial Intelligence}, vol. 101, p. 104171, 2021.

\bibitem{zhuang2022underwater}
P.~Zhuang, J.~Wu, F.~Porikli, and C.~Li, ``Underwater image enhancement with hyper-laplacian reflectance priors,'' \emph{IEEE Transactions on Image Processing}, vol.~31, pp. 5442--5455, 2022.

\bibitem{ancuti2017color}
C.~O. Ancuti, C.~Ancuti, C.~De~Vleeschouwer, and P.~Bekaert, ``Color balance and fusion for underwater image enhancement,'' \emph{IEEE Transactions on Image Processing}, vol.~27, no.~1, pp. 379--393, 2017.

\bibitem{kang2022perception}
Y.~Kang, Q.~Jiang, C.~Li, W.~Ren, H.~Liu, and P.~Wang, ``A perception-aware decomposition and fusion framework for underwater image enhancement,'' \emph{IEEE Transactions on Circuits and Systems for Video Technology}, vol.~33, no.~3, pp. 988--1002, 2022.

\bibitem{wang2019underwater}
K.~Wang, Y.~Hu, J.~Chen, X.~Wu, X.~Zhao, and Y.~Li, ``Underwater image restoration based on a parallel convolutional neural network,'' \emph{Remote sensing}, vol.~11, no.~13, p. 1591, 2019.

\bibitem{kar2021zero}
A.~Kar, S.~K. Dhara, D.~Sen, and P.~K. Biswas, ``Zero-shot single image restoration through controlled perturbation of koschmieder's model,'' in \emph{Computer Vision and Pattern Recognition}, 2021, pp. 16\,205--16\,215.

\bibitem{li2020underwater}
C.~Li, S.~Anwar, and F.~Porikli, ``Underwater scene prior inspired deep underwater image and video enhancement,'' \emph{Pattern Recognition}, vol.~98, p. 107038, 2020.

\bibitem{li2022beyond}
K.~Li, L.~Wu, Q.~Qi, W.~Liu, X.~Gao, L.~Zhou, and D.~Song, ``Beyond single reference for training: underwater image enhancement via comparative learning,'' \emph{IEEE Transactions on Circuits and Systems for Video Technology}, 2022.

\bibitem{naik2021shallow}
A.~Naik, A.~Swarnakar, and K.~Mittal, ``Shallow-uwnet: Compressed model for underwater image enhancement (student abstract),'' in \emph{AAAI Conference on Artificial Intelligence}, vol.~35, 2021, pp. 15\,853--15\,854.

\bibitem{fu2022uncertainty}
Z.~Fu, W.~Wang, Y.~Huang, X.~Ding, and K.-K. Ma, ``Uncertainty inspired underwater image enhancement,'' in \emph{European Conference on Computer Vision}, 2022, pp. 465--482.

\bibitem{wang2022semantic}
D.~Wang, L.~Ma, R.~Liu, and X.~Fan, ``Semantic-aware texture-structure feature collaboration for underwater image enhancement,'' in \emph{IEEE International Conference on Robotics and Automation}, 2022, pp. 4592--4598.

\bibitem{sharma2023wavelength}
P.~Sharma, I.~Bisht, and A.~Sur, ``Wavelength-based attributed deep neural network for underwater image restoration,'' \emph{ACM Transactions on Multimedia Computing Communications and Applications}, vol.~19, no.~1, pp. 1--23, 2023.

\bibitem{guo2023underwater}
C.~Guo, R.~Wu, X.~Jin, L.~Han, W.~Zhang, Z.~Chai, and C.~Li, ``Underwater ranker: Learn which is better and how to be better,'' in \emph{{AAAI Conference on Artificial Intelligence}}, vol.~37, no.~1, 2023, pp. 702--709.

\bibitem{zhao2024toward}
C.~Zhao, W.~Cai, C.~Dong, and Z.~Zeng, ``Toward sufficient spatial-frequency interaction for gradient-aware underwater image enhancement,'' in \emph{IEEE International Conference on Acoustics, Speech, and Signal Processing}, 2024, pp. 3220--3224.

\bibitem{zhu2017unpaired}
J.-Y. Zhu, T.~Park, P.~Isola, and A.~A. Efros, ``Unpaired image-to-image translation using cycle-consistent adversarial networks,'' in \emph{International Conference on Computer Vision}, 2017, pp. 2223--2232.

\bibitem{fabbri2018enhancing}
C.~Fabbri, M.~J. Islam, and J.~Sattar, ``Enhancing underwater imagery using generative adversarial networks,'' in \emph{IEEE International Conference on Robotics and Automation}, 2018, pp. 7159--7165.

\bibitem{han2021single}
J.~Han, M.~Shoeiby, T.~Malthus, E.~Botha, J.~Anstee, S.~Anwar, R.~Wei, L.~Petersson, and M.~A. Armin, ``Single underwater image restoration by contrastive learning,'' in \emph{IEEE International Geoscience and Remote Sensing Symposium}, 2021, pp. 2385--2388.

\bibitem{liu2022twin}
R.~Liu, Z.~Jiang, S.~Yang, and X.~Fan, ``Twin adversarial contrastive learning for underwater image enhancement and beyond,'' \emph{IEEE Transactions on Image Processing}, vol.~31, pp. 4922--4936, 2022.

\bibitem{ma2022wavelet}
Z.~Ma and C.~Oh, ``A wavelet-based dual-stream network for underwater image enhancement,'' in \emph{IEEE International Conference on Acoustics, Speech, and Signal Processing}, 2022, pp. 2769--2773.

\bibitem{cong2023pugan}
R.~Cong, W.~Yang, W.~Zhang, C.~Li, C.-L. Guo, Q.~Huang, and S.~Kwong, ``Pugan: Physical model-guided underwater image enhancement using gan with dual-discriminators,'' \emph{IEEE Transactions on Image Processing}, 2023.

\bibitem{ren2022reinforced}
T.~Ren, H.~Xu, G.~Jiang, M.~Yu, X.~Zhang, B.~Wang, and T.~Luo, ``Reinforced swin-convs transformer for simultaneous underwater sensing scene image enhancement and super-resolution,'' \emph{IEEE Transactions on Geoscience and Remote Sensing}, vol.~60, pp. 1--16, 2022.

\bibitem{fu2022unsupervised}
Z.~Fu, H.~Lin, Y.~Yang, S.~Chai, L.~Sun, Y.~Huang, and X.~Ding, ``Unsupervised underwater image restoration: From a homology perspective,'' in \emph{AAAI Conference on Artificial Intelligence}, vol.~36, 2022, pp. 643--651.

\bibitem{qi2023deep}
H.~Qi, H.~Zhou, J.~Dong, and X.~Dong, ``Deep color-corrected multi-scale retinex network for underwater image enhancement,'' \emph{IEEE Transactions on Geoscience and Remote Sensing}, 2023.

\bibitem{li2017watergan}
J.~Li, K.~A. Skinner, R.~M. Eustice, and M.~Johnson-Roberson, ``Watergan: Unsupervised generative network to enable real-time color correction of monocular underwater images,'' \emph{IEEE Robotics and Automation letters}, vol.~3, no.~1, pp. 387--394, 2017.

\bibitem{dietterich2002ensemble}
T.~G. Dietterich \emph{et~al.}, ``Ensemble learning,'' \emph{The handbook of brain theory and neural networks}, vol.~2, no.~1, pp. 110--125, 2002.

\bibitem{chen2022simple}
L.~Chen, X.~Chu, X.~Zhang, and J.~Sun, ``Simple baselines for image restoration,'' in \emph{European Conference on Computer Vision}, 2022, pp. 17--33.

\bibitem{zhu2018quaternion}
X.~Zhu, Y.~Xu, H.~Xu, and C.~Chen, ``Quaternion convolutional neural networks,'' in \emph{European Conference on Computer Vision}, 2018, pp. 631--647.

\bibitem{he2015spatial}
K.~He, X.~Zhang, S.~Ren, and J.~Sun, ``Spatial pyramid pooling in deep convolutional networks for visual recognition,'' \emph{IEEE transactions on pattern analysis and machine intelligence}, vol.~37, no.~9, pp. 1904--1916, 2015.

\bibitem{zamir2022restormer}
S.~W. Zamir, A.~Arora, S.~Khan, M.~Hayat, F.~S. Khan, and M.-H. Yang, ``Restormer: Efficient transformer for high-resolution image restoration,'' in \emph{Computer Vision and Pattern Recognition}, 2022, pp. 5728--5739.

\bibitem{mobley1993comparison}
C.~D. Mobley, B.~Gentili, H.~R. Gordon, Z.~Jin, G.~W. Kattawar, A.~Morel, P.~Reinersman, K.~Stamnes, and R.~H. Stavn, ``Comparison of numerical models for computing underwater light fields,'' \emph{Applied Optics}, vol.~32, no.~36, pp. 7484--7504, 1993.

\bibitem{wang2004image}
Z.~Wang, A.~C. Bovik, H.~R. Sheikh, and E.~P. Simoncelli, ``Image quality assessment: from error visibility to structural similarity,'' \emph{IEEE Transactions on Image Processing}, vol.~13, no.~4, pp. 600--612, 2004.

\bibitem{zhang2018unreasonable}
R.~Zhang, P.~Isola, A.~A. Efros, E.~Shechtman, and O.~Wang, ``The unreasonable effectiveness of deep features as a perceptual metric,'' in \emph{Computer Vision and Pattern Recognition}, 2018, pp. 586--595.

\bibitem{yang2015underwater}
M.~Yang and A.~Sowmya, ``An underwater color image quality evaluation metric,'' \emph{IEEE Transactions on Image Processing}, vol.~24, no.~12, pp. 6062--6071, 2015.

\bibitem{panetta2015human}
K.~Panetta, C.~Gao, and S.~Agaian, ``Human-visual-system-inspired underwater image quality measures,'' \emph{IEEE Journal of Oceanic Engineering}, vol.~41, no.~3, pp. 541--551, 2015.

\bibitem{guo2022underwater}
P.~Guo, H.~Liu, D.~Zeng, T.~Xiang, L.~Li, and K.~Gu, ``An underwater image quality assessment metric,'' \emph{IEEE Transactions on Multimedia}, vol.~25, pp. 5093--5106, 2022.

\bibitem{liu2023uiqi}
Y.~Liu, K.~Gu, J.~Cao, S.~Wang, G.~Zhai, J.~Dong, and S.~Kwong, ``Uiqi: A comprehensive quality evaluation index for underwater images,'' \emph{IEEE Transactions on Multimedia}, 2023.

\end{thebibliography}


\begin{IEEEbiography}[{\includegraphics[width=1in,height=1.25in,clip,keepaspectratio]{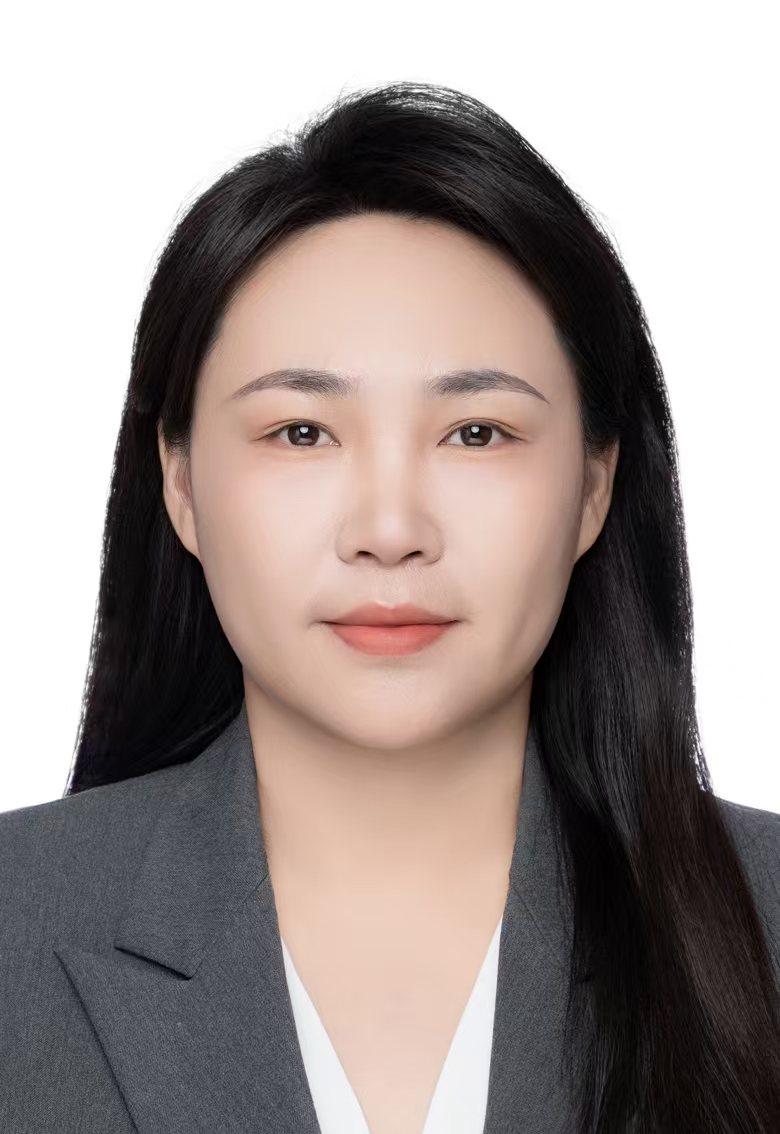}}]{Xiaojiao Guo}
received both the B.Eng. degree in software engineering and M.Eng. degree in software engineering from Sun Yat-sen University, Guangzhou, China, in 2009 and 2012 respectively. She is currently pursuing her Ph.D. degree with the Department of Computer and Information Science at the University of Macau, Macao, China. She is also a lecturer with the Baoshan Univeristy, Yunnan, China. Her current research interests include low-level computer vision tasks.
\end{IEEEbiography}

\begin{IEEEbiography}[{\includegraphics[width=1in,height=1.25in,clip,keepaspectratio]{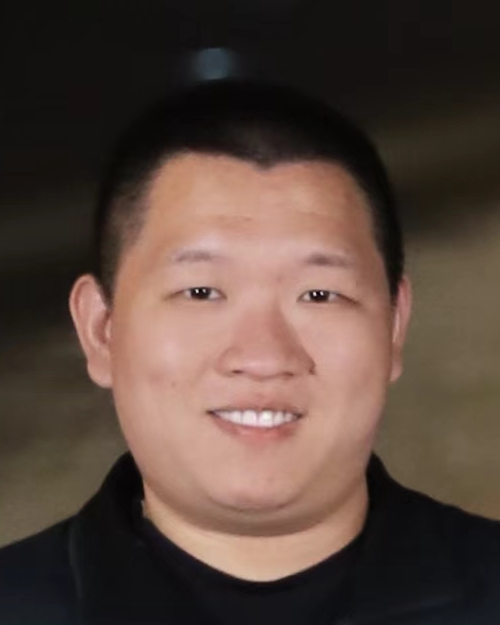}}]{Xuhang Chen} received the B.Sc. degree in electronic information science and technology from the Sun Yat-Sen University, Guangzhou, China, in 2016 and B.Eng. degree in electronic engineering from the Chinese University of Hong Kong, Hong Kong, China, in 2016, and the M.Eng. degree in electrical engineering and the M.Sc. degree in computer and information technology from the University of Pennsylvania, Philadelphia, USA, in 2019. He is currently pursuing his Ph.D. degree with the Department of Computer and Information Science, University of Macau, Macao, China. He is also a lecturer with the Huizhou Univeristy, Huizhou, China. His current research interests include computational photography and artificial intelligence.
\end{IEEEbiography}

\begin{IEEEbiography}[{\includegraphics[width=1in,height=1.25in,clip,keepaspectratio]{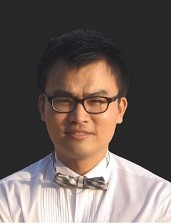}}]{Shuqiang Wang}
(SM'20) received his Ph.D. degree from City University of Hong Kong. He was a research scientist of Huawei Technologies Noah's Ark Lab. He held a postdoctoral fellowship at the University of Hong Kong from 2013 until 2014. He is currently a professor with Shenzhen Institutes of Advanced Technology (SIAT), Chinese Academy of Science. He has published more than 150 papers on journals, such as IEEE TPAMI/TNNLS/TIP/TSP/TMI etc. His current research interests include machine learning, brain image computing and optimization theory.
\end{IEEEbiography}

\begin{IEEEbiography}[{\includegraphics[width=1in,height=1.25in,clip,keepaspectratio]{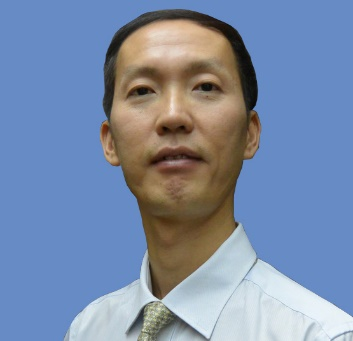}}]{Chi-Man Pun}
(Senior Member, IEEE) received his Ph.D. degree in Computer Science and Engineering from the Chinese University of Hong Kong in 2002, and his M.Sc. and B.Sc. degrees from the University of Macau. He had served as the Head of the Department of Computer and Information Science, University of Macau from 2014 to 2019, where he is currently a Professor and in charge of the Image Processing and Pattern Recognition Laboratory. He has investigated many externally funded research projects as PI, and has authored/co-authored more than 200 refereed papers in many top-tier journals and conferences. He has also co-invented several China/US Patents, and is the recipient of the Macao Science and Technology Award 2014 and the Best Paper Award in the 6th Chinese Conference on Pattern Recognition and Computer Vision (PRCV2023). Dr. Pun has served as the General Chair/Co-Chair/Program Chair for many international conferences such as the 10th and 11th International Conference on Computer Graphics, Imaging and Visualization (CGIV2013, CGIV2014), the IEEE International Conference on Visual Communications and Image Processing (VCIP2020) and the International Workshop on Advanced Image Technology (IWAIT2022), etc., and served as the SPC/PC member for many top-tier conferences such as AAAI, IJCAI, CVPR, ICCV, ECCV, MM, etc. He is currently serving as the Editorial Board member for the journal of Artificial Intelligence (AIJ). Besides, he has been listed in the World's Top 2\% Scientists by Stanford University since 2020. His research interests include Image Processing and Pattern Recognition; Multimedia and AI Security; Medical Image Analysis, etc.
\end{IEEEbiography}

 





\end{document}